%% file: main.tex
\newcommand{\CLASSINPUTtoptextmargin}{0.8in}
\newcommand{\CLASSINPUTbottomtextmargin}{0.97in}
\documentclass[journal]{IEEEtran}
\IEEEoverridecommandlockouts

\usepackage{graphicx} 

\usepackage{mathtools}
\DeclarePairedDelimiter\ceil{\lceil}{\rceil}

\usepackage{hyperref}
\usepackage{dsfont}
\usepackage{cite}
\usepackage{amsmath,amssymb,amsfonts}
\usepackage{nicefrac}
\usepackage{pifont}
\usepackage{comment}

\usepackage{color,soul}

\usepackage[ruled,linesnumbered]{algorithm2e}

\usepackage{algorithmic}

\usepackage{graphicx}
\usepackage{textcomp}
\usepackage{xcolor}

\usepackage{mathtools, stmaryrd}
\usepackage{xparse} 
\usepackage{mathtools, nccmath,amsmath,amssymb,amsfonts,amstext, dsfont, color}
\usepackage{subcaption}
\usepackage{float}
\usepackage{booktabs}

\usepackage{array,multirow}

\usepackage{xcolor}

\definecolor{lightgray}{gray}{0.9}

\usepackage{mdframed}

\newmdtheoremenv[
  linecolor=black,
  backgroundcolor=lightgray,
  linewidth=2pt,
  innerleftmargin=5pt,
  innerrightmargin=5pt,
  innertopmargin=5pt,
  innerbottommargin=5pt
]{theoremSp}{Theorem}

\newmdtheoremenv[
  linecolor=black,
  backgroundcolor=lightgray,
  linewidth=2pt,
  innerleftmargin=5pt,
  innerrightmargin=5pt,
  innertopmargin=5pt,
  innerbottommargin=5pt
]{lemmaSp}{Lemma}

\newmdtheoremenv[
  linecolor=black,
  backgroundcolor=lightgray,
  linewidth=2pt,
  innerleftmargin=5pt,
  innerrightmargin=5pt,
  innertopmargin=5pt,
  innerbottommargin=5pt
]{corollarySp}{Corollary}

\newtheorem{assumption}{Assumption}

\newcommand{\Proba}[1]{\mathrm{Pr}\left(#1\right)}
\newcommand{\E}[1]{\mathds{E}\left[#1\right]}
\newcommand{\Ex}[2]{\mathds{E}_{#2}\left[#1\right]}

\newcommand{\norm}[1]{\lVert #1 \rVert}
\newcommand{\scalar}[2]{\langle #1 , #2  \rangle}

\newcommand{\hatbf}[1]{\hat{\mathbf{#1}}}

\newcommand{\bm}[1]{\mathbf{#1}}
\DeclareMathOperator*{\argmax}{arg\,max}
\DeclareMathOperator*{\argmin}{arg\,min}

\begin{document}
\pagestyle{plain}
\title{Efficient and Optimal No-Regret Caching under Partial Observation}

\author{%
  Younes Ben Mazziane\textsuperscript{1}, 
  Francescomaria Faticanti\textsuperscript{2}, 
  Sara Alouf\textsuperscript{3}, 
  and Giovanni Neglia\textsuperscript{3}\\
  \textsuperscript{1}\footnotesize Avignon University, Avignon, France. Email: younes.ben-mazziane@univ-avignon.fr\\
  \textsuperscript{2}\footnotesize ENS Lyon, Lyon, France. Email: francescomaria.faticanti@ens-lyon.fr\\
  \textsuperscript{3}\footnotesize Inria, Université Côte d'Azur, Sophia Antipolis, France. Emails: \{sara.alouf, giovanni.neglia\}@inria.fr
}

\maketitle

\begin{abstract}
\input{Journal/abstract}

\end{abstract}

\begin{IEEEkeywords}
Caching, Online learning, Follow-the-Perturbed-Leader. 
\end{IEEEkeywords}

\section{Introduction}\label{sec:intro}
\input{Journal/introYB}

\section{Problem Formulation} 
\label{s:Caching-problem}
\input{Journal/problem}

\section{Background and Related Work}\label{sec:bacground}
\input{Journal/background}

\section{{The NFPL caching policy}}\label{sec:extending}

\input{Journal/approach}

\section{Numerical Evaluation}\label{sec:experiments}
\input{Journal/experiments}


\section{Conclusion}\label{sec:conclusions}

\input{Journal/conclusions}

\section{Acknowledgments}
This research was supported in part by the French Government through the “Plan de Relance” and “Programme d’Investissements d’Avenir”.

\bibliography{main.bib}
\bibliographystyle{ieeetr}


\newpage

\section{Supplementary Material}
\label{sec:ProofThMain}

\input{Journal/ProofMainTh}



\end{document}

%% file: Journal/abstract.tex
Online learning algorithms have been successfully used to design caching policies with sublinear regret in the total number of requests, with no statistical assumption about the request sequence. Most existing algorithms involve computationally expensive operations and require knowledge of all past requests. However, this may not be feasible in practical scenarios like caching at a cellular base station. Therefore, we study the caching problem in a more restrictive setting where only a fraction of past requests are observed, and we propose a randomized caching policy with sublinear regret based on the classic online learning algorithm Follow-the-Perturbed-Leader (FPL). Our caching policy is the first to attain the asymptotically optimal regret bound while ensuring asymptotically constant amortized time complexity in the partial observability setting of requests. The experimental evaluation compares the proposed solution against classic caching policies and validates the proposed approach under synthetic and real-world request traces.

%% file: Journal/introYB.tex
Caching techniques are extensively employed in computer systems, serving various purposes such as accelerating CPU performance~\cite{tse1998cpu} and enhancing user experiences in content delivery networks (CDNs)~\cite{bektas2007designing}. The primary objective of a caching system is to carefully choose files for storage in the cache to maximize the proportion of file requests that can be fulfilled locally. This approach effectively minimizes the dependence on remote server retrievals, which can be costly in terms of delay and network traffic.
The presence of caching systems facilitates more efficient data delivery 
and leads to enhanced overall system performance, especially with the widespread adoption of traffic-intensive applications such as virtual and augmented reality~\cite{chatzopoulos2017mobile}, or edge video analytics~\cite{ananthanarayanan2017real}.

Caching policies have been thoroughly investigated under varied assumptions concerning the statistical regularity of file request processes~\cite{IRM,traverso2013temporal}. However, real-world request sequences tend to deviate from these theoretical models, especially when aggregated over small geographic areas~\cite{leconte2016placing}. This deviation has inspired the exploration of online learning algorithms, beginning with the work of Paschos et al.~\cite{paschos2019learning}, who proposed the Online Gradient Descent (\texttt{OGD}) algorithm~\cite{zinkevich2003online} for caching. Online learning algorithms exhibit robustness to varying request process patterns, as they operate under the assumption that requests may be generated by an adversary.

In this context, the main metric of interest is the \emph{regret}, which is the difference between the cost---e.g., the expected number of cache misses---incurred by a given online caching algorithm
and the cost of the optimal static cache allocation with hindsight, i.e., with knowledge of the future requests over a fixed number of requests. In this framework, the primary objective is to design \emph{no-regret} algorithms, i.e., online policies whose regret grows sublinearly with the total number of requests~\cite{paschos2019learning}.

Several no-regret caching policies have been proposed in the literature, drawing on well-known online learning algorithms such as Online Gradient Descent (\texttt{OGD})~\cite{paschos2019learning}, Online Mirror Descent (\texttt{OMD})~\cite{si2023no}, Follow-the-Regularized-Leader (\texttt{FRL})~\cite{mhaisen2022online}, and Follow-the-Perturbed-Leader (\texttt{FPL})~\cite{bhattacharjee2020fundamental}. However, most no-regret policies rely on computationally expensive operations and assume access to the complete history of past requests. In practice, many scenarios require caching policies to operate under \emph{partial observations} of the request process.

For instance, in femtocaching systems~\cite{FEMTOCaching13TIT}, a central base station manages the content of multiple caches located at smaller cell base stations, known as \textit{helpers}. While helpers can independently select their cache content, performing computationally intensive tasks, such as those required by no-regret algorithms, often exceeds their processing capabilities. Consequently, such operations may need to be offloaded to the base station, which has substantially greater computational resources. However, under standard conditions, the base station only has visibility into cache misses at the helpers. Gaining complete knowledge of past file requests would necessitate continuous communication between the helpers and the base station, thereby introducing significant overhead.

Additionally, partial observations can also result from request routing mechanisms in CDNs. For example, requests are often routed---via DNS redirection or similar techniques---to caches believed to store the requested items, thereby further restricting the caching policy's visibility into the complete request process. 

To the best of our knowledge, only \cite{BuraPartial_Obser_caching_TON22} and \cite{liu2022learning} have explored caching policies under a specific partial observation regime, where the cache is only aware of requests for items it stores. The former focuses exclusively on scenarios where requests follow a stationary stochastic process. The latter avoids making statistical assumptions about the request process but proposes a policy with an amortized time complexity that is quadratic in the catalog size and a regret bound that scales linearly with the same parameter.

\subsection{Contributions}

This paper investigates caching under a partial observability framework, which we refer to as the Bernoulli Partial Observability (BPO) regime, where the caching policy observes each request---whether for cached or noncached files---with probability~$p$. 
In the context of the two motivating examples discussed earlier, the probability~$p$ corresponds to the fraction of requests forwarded from the helpers to the base station in the femtocaching scenario, and the fraction of CDN requests routed to a specific cache in the CDN setting.




\textit{Our primary contribution is proving that simple modifications to the \texttt{FPL} caching policy yield asymptotically optimal regret bounds in the cache size, the total number of files, and the trace length, and scales as $\nicefrac{1}{p}$, in the more restrictive BPO regime. Moreover, this new policy, which we call \texttt{NFPL}, has parameters that enable a trade-off among the expected cost (measured by the regret metric), the variability around this expected cost, and the amortized time complexity. In particular, with a specific configuration of these parameters, \texttt{NFPL} is the first no-regret caching policy with an amortized time complexity of~$\mathcal{O}(1)$ as the total number of requests approaches infinity.}


The regret bound of \texttt{NFPL} is presented in Theorem~\ref{th:RegretNFPLMain}. To prove this result, we study an online optimization framework where noisy estimates of linear loss functions are revealed sequentially, e.g., noise is due to partial observability of the loss functions. The goal is to dynamically adjust decisions to minimize the cumulative sum of the actual loss functions. We establish sufficient conditions under which the extension of \texttt{NFPL} to this more restrictive setting achieves sublinear regret. In particular, these conditions are met by \texttt{NFPL} in the BPO caching model, confirming its no-regret property in that context. 

The \texttt{NFPL} policy, which is described in Alg.~\ref{algo:NFPL-singleCache}, closely resembles existing \texttt{FPL} caching approaches in the literature~\cite{bhattacharjee2020fundamental,BuraPartial_Obser_caching_TON22,liu2022learning}. Specifically, it leverages request counts and randomly generated vectors to guide its decisions. This allows \texttt{NFPL} to cache relevant files that a greedy policy, such as Least-Frequently-Used (\texttt{LFU}), might overlook. However, \texttt{NFPL} introduces new features to balance regret and time complexity. It utilizes a batching approach, where cache updates occur only after collecting a batch of requests, similarly to some previously proposed no-regret policies~\cite{si2023no,FaticantiFRL24}. Furthermore, \texttt{NFPL} intentionally omits requests with a specified probability to decrease the frequency of cache updates. It also allows for controlled temporal correlations of its random vectors. In particular, we show that a specific coupling of these noise vectors, inspired by a variant of \texttt{FPL} called Follow-The-Lazy-Leader~\cite{kalai2005efficient}, leads to an efficient implementation that infrequently changes the cache state. Indeed, we bound the probability of such changes, demonstrating that \texttt{NFPL} is the first no-regret caching policy with $\mathcal{O}(1)$ amortized time complexity when the total number of requests approaches infinity. This result is presented in Theorem~\ref{th:TimeComplexityNFPL}.


We complement the theoretical findings on the regret and time complexity of \texttt{NFPL} with simulations on two synthetic traces with Zipf-distributed popularities---one has stationary content popularity patterns and the other is adversarial---and a real-world trace from Akamai~\cite{neglia2017access}. Simulations evaluate the performance of \texttt{NFPL} in the BPO regime in terms of the average miss ratio, its variance, and execution time. Notably, the \texttt{NFPL} caching policy performs exceptionally well on the adversarial trace, significantly outperforming alternative approaches. These numerical results underscore \texttt{NFPL}'s robustness and adaptability, even in the presence of non-stationary request sequences.

This paper extends our previous work~\cite{mazziane2023no} by showing that a careful choice of correlations between the noise vectors in \texttt{NFPL} leads to $\mathcal{O}(1)$ amortized time complexity while maintaining the regret guarantees. Furthermore, the numerical results on the adversarial trace with skewed file popularity, highlighting the robustness of \texttt{NFPL}, are new.

\subsection{Road map}

The rest of the paper is organized as follows: Section~\ref{s:Caching-problem} formally describes the caching problem tackled in this paper, and Section~\ref{sec:bacground} provides background details and discusses related work. The \texttt{NFPL} algorithm and its regret and time complexity guarantees are presented in Section~\ref{sec:extending}. A numerical evaluation of our approach is shown in Section~\ref{sec:experiments}. Finally, Section~\ref{sec:conclusions} concludes the paper. Detailed proofs and additional numerical results are available in the supplementary material.

%% file: Journal/problem.tex
In what follows, the notation~$[n]$ designates the set~$\{1, \ldots ,n\}$ for any positive integer~$n$. As commonly used, $\mathds{1}(F)$ stands for the indicator function that~$F$ is true. We use bold notation to represent vectors and matrices. 

We consider a server storing a set~$\mathcal{I}$ of~$N$ files and a single cache memory, with limited computational capabilities, that can store up to~$C<N$ files from~$\mathcal{I}$. A sequence of requests for items in~$\mathcal{I}$ of length~$T$, denoted~$\mathbf{f} = (f_s)_{s\in [T]}$, arrives at the cache. If~$f_t$ is available in the cache at step~$t$, the request is a \textit{hit}, and the cache serves the request. Otherwise, it is a \textit{miss}, and the cache forwards the request to the server.

\begin{table}
\small
  \caption{Table of notation.}
\label{tab:notation}
  \centering
\begin{tabular}{ |l|l| } 
 \multicolumn{2}{c}{{\bf Caching problem}} \\
 \hline
 $\mathcal{I}$ & set of files \\
 $N=|\mathcal{I}|$ & catalog size \\
 $C$ & cache capacity \\
$T$ & total number of requests  \\ 
$\mathbf{f}=(f_t)_{t\in [T]}$ & sequence of requests \\
$S_{\mathcal{A}}(t)$ & files stored by the policy $\mathcal{A}$ \\
$\delta_t$ & observability of the request by~$\mathcal{A}$  \\ 
$p$ & success probability of $(\delta_t)_{t\in [T]}$\\
$\mathcal{R}_T(\mathcal{A})$ & regret of policy $\mathcal{A}$ \\
 \hline
 \multicolumn{2}{c}{ \textbf{NFPL}} \\
\hline
$\beta_t$ & sampling the request at step $t$ \\ 
$q$ & success probability of $(\beta_t)_{t\in [T]}$ \\ 
$B$ & batch size in \texttt{NFPL}\\ 
$\boldsymbol{\gamma}_t= (\gamma_{t,f})_{f\in\mathcal{I}}$& randomly generated noise vectors \\
$\eta$ & parameter of $(\boldsymbol{\gamma}_t)_{t\in [T]}$ \\
\hline
 \multicolumn{2}{c}{{\bf Online linear learning}} \\ 
 \hline
$\mathcal{X}$ & decision set \\
$\mathcal{B}$ & costs set \\
$N$ & dimension of vectors in $\mathcal{X}$ and $\mathcal{B}$\\
$T$ & time horizon \\
$\mathbf{r}_t$ & cost vector at step $t$ \\
$\mathbf{x}_t$ & decision vector  \\
$\langle \mathbf{r}_t,\mathbf{x}_t \rangle$ & cost paid \\
$\mathbf{r}_{1:t}$ & sum of $\mathbf{r}_s$ for all values of $s$ from $1$ to $t$ \\
$M(\mathbf{r})$ & value of $\mathbf{x}$ in $\mathcal{X}$ that minimizes $\langle \mathbf{r},\mathbf{x} \rangle$ \\
$\mathcal{R}_T(\mathcal{A})$ & regret of algorithm~$\mathcal{A}$ \\ 
$\hat{\mathbf{r}}_t$ & noisy cost vectors estimates \\
$\hat{\mathcal{B}}$ & state space of $(\hat{\mathbf{r}}_t)_{t\in [T]}$ \\
\hline
\end{tabular}  
\end{table}

A caching policy~$\mathcal{A}$ decides the content of the cache at time~$t$, denoted~$S_{\mathcal{A}}(t)$, based on past requests~$f_1,\ldots ,f_{t-1}$. Algorithm~$\mathcal{A}$ samples~$S_{\mathcal{A}}(t)$ from a probability distribution~$\mathcal{P}_{\mathcal{A}}(t)$ over the space~$\{S\subset \mathcal{I}: |S|=C\}$, which we denote as~$\binom{\mathcal{I}}{C}$. We adopt the \textit{oblivious} adversarial model, wherein the request sequence can be arbitrary but must be independent of the random choices $(S_{\mathcal{A}}(t))_t$. In this model, the request sequence of fixed length $T$ can be sampled from an arbitrary probability distribution, thus covering typical stationary and non-stationary stochastic models. A common performance metric for an algorithm~$\mathcal{A}$ under adversarial models is its regret, denoted as~$\mathcal{R}_T(\mathcal{A})$, and defined as follows, 
\begin{align}\label{e:defRegretCaching}
            \mathcal{R}_{T}(\mathcal{A}) \triangleq \sup_{\mathbf{f} \in \mathcal{I}^{T}} \big(  \Ex{M_{\mathcal{A}}(\mathbf{f})}{\boldsymbol{\mathcal{P}}_{\mathcal{A}}} - M^{*}(\mathbf{f}) \big),
\end{align}
where~$\boldsymbol{\mathcal{P}}_{\mathcal{A}}= (\mathcal{P}_{\mathcal{A}}(t))_{t\in [T]}$, $M_{\mathcal{A}}(\mathbf{f}) \triangleq  \sum_{t=1}^{T} \mathds{1}\left(f_t \notin S_{\mathcal{A}}(t) \right)$ is the number of misses of~$\mathcal{A}$, 
and $M^{*}(\mathbf{f})= \min_{S \in \binom{\mathcal{I}}{C}} \sum_{t=1}^{T} \mathds{1}\left(f_t \notin S \right)$ is the number of misses of the optimal static caching policy with knowledge of~$\mathbf{f}$. 
The purpose is to design a no-regret policy---one whose regret grows sublinearly in~$T$.  



While many no-regret caching policies have been proposed recently, most require the caching policy~$\mathcal{A}$ to be aware of all past requests for the cache. We consider a more restrictive framework, where the caching policy~$\mathcal{A}$ has access to a \textit{subset of requests}. We model the availability of the request at step~$t$ for~$\mathcal{A}$ via a Bernoulli random variable~$\delta_t$. If~$\delta_t = 1$, it indicates that~$\mathcal{A}$ is aware of the request at that particular step. We assume that $(\delta_t)_{t\in [T]}$ are independent and identically distributed (i.i.d.) with success probability denoted~$p$. This framework is referred to as the \textit{Bernoulli Partial Observability} (BPO) regime.


Designing a no-regret caching policy under the BPO regime enables achieving sublinear regret in two related partial observability settings, which we term Hit Partial Observability (HPO) and Miss Partial Observability (MPO). In the MPO regime, requests for cached files are always observed, while requests for non-cached files are observed with probability $p$. Formally, $\Proba{\delta_t = 1 | f_t \in S_{\mathcal{A}}(t-1)} = 1$, and $\Proba{\delta_t = 1 | f_t \notin S_{\mathcal{A}}(t-1)} = p$. The HPO regime mirrors this structure, but with the roles of cached and non-cached files reversed. By deliberately sampling requests for cached files with probability~$p$ in the MPO regime, the observation process becomes equivalent to that of the BPO regime. Consequently, any no-regret caching policy designed for BPO can be adapted directly to HPO and MPO.



In addition to the no-regret property, it is important to take into account the computational cost of the algorithm that selects which files to cache based on the observed requests. To this end, we employ the amortized time complexity over the~$T$ rounds of~$\mathcal{A}$, i.e., the average time complexity of $\mathcal{A}$ per request, as a performance metric.

%% file: Journal/background.tex



    




   
    


\subsection{Online linear learning and FPL}
\label{ss:OnlineLearning_FPL}
In the online linear learning setting, an agent sequentially makes decisions in~$T$ rounds. At each round~$t$, the agent selects a decision vector~$\mathbf{x}_t$ from a set~$\mathcal{X}\subset \mathbb{R}^{N}$ and then observes a cost vector $\mathbf{r}_t$ from a set $\mathcal{B}\subset \mathbb{R}^{N}$. The agent then incurs a loss $\langle \mathbf{x}_t, \mathbf{r}_t \rangle$, where $\scalar{\mathbf{r}}{\mathbf{x}} \triangleq \sum_{i=1}^N r_{i} x_i$ denotes the scalar product of the two vectors $\mathbf{r}$ and $\mathbf{x}$. In this context, the metric of performance for an algorithm $\mathcal{A}$ selecting a decision vector $\mathbf{x}_{t} = {\mathcal{A}}(\bm{r}_1, \ldots, \bm{r}_{t-1})$ is the regret defined as follows,

\begin{equation}
    \mathcal{R}_T(\mathcal A) = \sup_{\{\mathbf{r}_1,\ldots,\mathbf{r}_T\}} \left \{ \E{\sum_{t=1}^{T} \scalar{\mathbf{r}_t}{\mathbf{x}_t}} - \min_{\mathbf{x}\in \mathcal{X}} \langle \mathbf{x}, \sum_{t=1}^{T} \mathbf{r}_t\rangle\right \}\; .
\label{eq:regret}
\end{equation} 
The objective is to design an algorithm with sublinear regret, $\mathcal{R}_T(\mathcal{A}) = o(T)$. These algorithms are commonly known as \textit{no-regret} algorithms since their time-average cost approaches the optimal static policy cost as $T$ grows.


An intuitive solution to minimize the regret, known as Follow-The-Leader (\texttt{FTL})~\cite{de2014follow}, is to greedily select the state that minimizes the past cumulative cost, i.e., $x_{t+1} = M(\mathbf{r}_{1:t})$, where $M(\mathbf{r})$ denotes an arbitrary element of $\arg\min_{\mathbf{x}\in \mathcal{X}} \scalar{\mathbf{r}}{\mathbf{x}}$ for any vector~$\mathbf{r}$, and $\mathbf{r}_{1:t}\triangleq \sum_{s=1}^t \mathbf{r}_s$ represents the aggregate sum of a given sequence of vectors $(\mathbf{r}_1, \dots, \mathbf{r}_t)$. Although \texttt{FTL} proves optimal when $(\bm{r}_t)_t$ are sampled from a stationary distribution, it unfortunately yields linear regret in adversarial settings~\cite{de2014follow}.

Follow-the-Perturbed Leader (\texttt{FPL}) improves the performance of \texttt{FTL} by incorporating, at each time step $t$, a noise vector $\boldsymbol{\gamma}_t$ of size $N$, whose components are i.i.d. random variables. Formally, 
\begin{equation}
    \label{e:fpl_update}
     \mathbf{x}_{t}(\texttt{FPL}) = M(\mathbf{r}_{1:t-1}+\boldsymbol{\gamma}_{t}) .
 \end{equation}
Let~$D\triangleq \sup_{\mathbf{x},\mathbf{y}\in \mathcal{X}} \norm{\mathbf{x}-\mathbf{y}}_1$ be the diameter of the decision set~$\mathcal{X}$, 
$A$ be a bound on the norm $1$ of vectors in the cost set $\mathcal{B}$, and $R$ be a bound on $\langle \mathbf{x},\mathbf{r} \rangle$ for any $(\mathbf{x},\mathbf{r})\in \mathcal{X}\times \mathcal{B}$. \cite[Thm. 1.1]{kalai2005efficient} shows that \texttt{FPL} is a no-regret algorithm for the broad class of online linear learning problems when the components of $\boldsymbol{\gamma}_t$ are uniform random variables. 

\begin{table*}[h]
\centering
\caption{Comparison of no-regret caching policies.}
\label{tab:time_regret_OCO_policies}
\begin{tabular}{|l|l|l|l|}
\hline
\textbf{Algorithm} & \textbf{Regime} & \textbf{Regret Bound} & \textbf{Time Complexity} \\ \hline
\texttt{FPL}\hfill~\cite{bhattacharjee2020fundamental,MukhopadhyaySSwitchingCost21}
    & BPO, $p=1$
    & $\mathcal{O}((\ln(N)^{1/4}) \sqrt{CT})$\quad\hfill\cite[Thm.~3]{bhattacharjee2020fundamental} 
    & $\mathcal{O}(\ln(C))$ \\ \hline
\texttt{OGD}\hfill~\cite{paschos2019learning}
    & BPO, $p=1$
    & $\mathcal{O}(\sqrt{CT})$~\hfill\cite[Thm.~1]{paschos2019learning} 
    & $\mathcal{O}(N)$ \\ \hline
\texttt{OMD} \hfill~\cite{si2023no}
    & BPO, $p=1$
    & $\mathcal{O}(C \sqrt{\ln(N) T})$~\hfill\cite[Cor.~2]{si2023no} 
    & $\mathcal{O}(N)$ \\ \hline
\texttt{FPLGR} \hfill~\cite{liu2022learning}
    & MPO, $p \in [0,1]$
    & $\mathcal{O}(NC \sqrt{T})$~\hfill\cite[Thm.~1]{liu2022learning} 
    & $\mathcal{O}(N^2 \ln(N))$ \\ \hline
\texttt{NFPL} ~\hfill[this paper]
    & BPO, $p \in (0,1]$  
    & $\mathcal{O}(\nicefrac{1}{p}\sqrt{CT})$ \hfill[Thm.~\ref{th:RegretNFPLMain}]
    & $\mathcal{O}(1)$ as $T\to +\infty$ ~[Thm.~\ref{th:TimeComplexityNFPL}] \\ \hline
\end{tabular}
\end{table*}

\begin{theoremSp}\cite[Thm. 1.1]{kalai2005efficient}
When $\boldsymbol{\gamma}_{t}$ is sampled from a multivariate uniform distribution with independent components over $[0, \eta]^N$, with $\eta = \sqrt{RAT/D}$, then:
\begin{align}
\mathcal{R}_T(\texttt{FPL}) \leq 2\sqrt{RADT}.
\end{align}
\end{theoremSp}

This no-regret property of \texttt{FPL} extends to other probability distributions for $\boldsymbol{\gamma}_t$, including exponential and Gaussian distributions~\cite{kalai2005efficient,abernethy2014online,CohenFPLStructruedLearning15}. These extensions may yield different regret bounds depending on the characteristics of the sets $\mathcal{X}$ and $\mathcal{B}$. Interestingly, \texttt{FPL} can be adapted to maintain its no-regret property in the \textit{multi-armed bandit} problem~\cite{Abernethy_noise_choice_LT15}, where decisions are represented by one-hot vectors and only the cost $\langle \bm{r}_t, \bm{x}_t \rangle$ is observed, rather than the cost vector $\bm{r}_t$.

In contrast to the online linear setting, where the decision set $\mathcal{X}$ could be arbitrary but the loss functions ($\langle \bm{r}_t, \cdot \rangle$) are linear, another well-studied framework considers the case where $\mathcal{X}$ is convex while allowing for more general convex loss functions. This setting falls within the domain of online convex optimization, where algorithms such as \texttt{OGD}, \texttt{OMD}, and \texttt{FRL} 
are known for their no-regret guarantees~\cite{hazan2016introduction}.


\subsection{Online learning for caching}

The caching problem naturally fits within the online linear learning framework; the set $\mathcal{X}$ models the possible cache allocations and the set $\mathcal{B}$ represents the requests. For example, one may define $x_{t,i} = 1$ if and only if $i \in S_{\mathcal{A}}(t)$, and $r_{t,i} = 1$ if and only if $f_t = i$; otherwise, the components of both vectors are zero. This formulation was considered in~\cite{bhattacharjee2020fundamental}, where the authors used \texttt{FPL} with Gaussian noise, following the spirit of~\cite{abernethy2014online,Abernethy_noise_choice_LT15}, but rather than directly adapting existing \texttt{FPL} results to the caching problem, they introduced a different proof technique. This approach allowed them to improve the dependence of the regret bound on $N$ and $C$ (see \cite[Thm. 3]{bhattacharjee2020fundamental}).

To leverage online convex optimization algorithms such as \texttt{OGD} and \texttt{OMD}, \cite{paschos2019learning,salem2021no} adopt an alternative modeling in which $x_{t,i}$ represents the fraction of the stored file, allowing for a convex decision set $\mathcal{X}$. In both works, the regret bounds and the algorithmic implementations had to be carefully adapted to account for the specific structure of caching. Furthermore, \cite{paschos2020online,PariaSLeadCache21} designed near-optimal regret algorithms for the more challenging problem of \textit{bipartite caching}.

\cite{liu2022learning} is the only work that considered the partial observability of adversarial requests. Specifically, they augmented \texttt{FPL} with a \textit{Geometric Sampling} procedure~\cite{FPLWithGR/jmlr/NeuB16}, to design a no-regret policy, that we call \texttt{FPLGR}, in the MPO regime with $p\in[0,1]$. 

Table~\ref{tab:time_regret_OCO_policies} reports the regime of observability of the requests, the regret bound, and the time complexity, for the above-mentioned caching policies, namely, \texttt{FPL}~\cite{bhattacharjee2020fundamental,MukhopadhyaySSwitchingCost21}, \texttt{OGD}~\cite{paschos2019learning}, \texttt{OMD}~\cite{si2023no}, \texttt{FPLGR}~\cite{liu2022learning}, and the policy of this paper \texttt{NFPL}. 

\noindent \textbf{NFPL}. Our policy operates in the BPO regime, which is more general than MPO when $p>0$. While our policy cannot operate in MPO with $p=0$, it leverages the additional fraction of observed information ($p \in (0,1]$) to achieve a time complexity of $\mathcal{O}(1)$ as $T\to \infty$, which is significantly better than the time complexity of existing caching policies, even when $p=1$. Furthermore, the regret of \texttt{NFPL} is optimal because \cite[Thm. 2]{bhattacharjee2020fundamental} proves a lower bound of~$\Omega(\sqrt{CT})$ on the regret of any caching policy.

\noindent \textbf{FPL and FPLGR.} The \texttt{FPL} caching policy achieves the best previously known time complexity of $\mathcal{O}(\ln{C})$ when $p=1$~\cite{bhattacharjee2020fundamental,MukhopadhyaySSwitchingCost21}.  \texttt{FPLGR} is an extension of \texttt{FPL} to maintain the no-regret property in the MPO regime. However, this 
comes at a significant cost: the regret bound scales linearly with $N$, and the time complexity increases to $\mathcal{O}(N^2 \ln N)$.

\noindent \textbf{OGD and OMD.} While \texttt{OGD} has an optimal regret bound, it is computationally expensive. \texttt{OMD} was proposed to mitigate this computational burden. In the fractional caching setting, performing cache updates after every time step yields update complexities of $\mathcal{O}(N)$ for \texttt{OGD} and $\mathcal{O}(C)$ for \texttt{OMD}. Conversely, if updates are performed only after collecting a batch of requests, the complexities become $\mathcal{O}(N^2)$ for \texttt{OGD} and $\mathcal{O}(N)$ for \texttt{OMD}. However, for the discrete caching problem addressed in this paper under the BPO setting with $p=1$, both \texttt{OGD} and \texttt{OMD} require an additional routine of complexity $\mathcal{O}(N)$ to convert a fractional cache state into a discrete one \cite{si2023no}. As a result, the overall time complexity for both algorithms is $\mathcal{O}(N)$. Recently, \cite{carra2024online} proposed a no-regret online gradient-based caching policy with a time complexity of $\mathcal{O}(\ln(N))$, achieved under a relaxed cache capacity constraint satisfied in expectation, i.e., $\E{|S_{\mathcal{A}}(t)|} = C$.

More generally, recent works have shown that online learning techniques can enhance edge caching performance. For example,~\cite{MhaisenSigmetrics,MhaisenIEEE_TMC24,FaticantiFRL24} propose optimistic no-regret algorithms that leverage neural network predictions about future requests, \cite{FanHM23infocomDynamicRegret} proposes a randomized caching algorithm with a low \textit{dynamic} regret, and \cite{Shroff_TON_24} uses online learning to estimate fetching costs.

%% file: Journal/approach.tex
    


No-regret caching policies typically rely on knowing the complete history of past requests. For instance, the \texttt{FPL} caching policy from~\cite{bhattacharjee2020fundamental} uses this knowledge to compute exact request counts for each file. However, this approach is incompatible with the BPO regime, formally defined in Section~\ref{s:Caching-problem}, where only approximate request counts are available. In response, we propose three \texttt{FPL} based caching policies with sublinear regret under the BPO regime. All these algorithms are derived from the same family that we call \texttt{NFPL}. The rest of this section is organized as follows: Section~\ref{ss:Description_NFPL} describes the dynamics of each \texttt{NFPL} algorithm, Section~\ref{ss:regret_guarantees} establishes the regret guarantees, Section~\ref{ss:time_complexity} analyzes the time complexity, and Section~\ref{ss:Tradeoff_regret_time} examines the trade-off between regret and computational efficiency.

\subsection{Policy description}
\label{ss:Description_NFPL}

\begin{algorithm}[t]
\caption{\texttt{NFPL}}
\begin{algorithmic}[1]
\STATE \textbf{Input:} Subsample of requests $\{f_t : \;t\in [T ], \delta_t=1\}$,
cache capacity~$C$, batch size $B$, sampling probability $q$
\STATE \textbf{Output:} $C$-sized set of stored content at each time step. 
\STATE  \( \hatbf{n}(0) = (\hat n_1(0), \ldots, \hat n_N(0)) )\gets \mathbf{0} \)
\STATE $\eta \gets \sqrt{\frac{BT}{2C}} $
\STATE $\boldsymbol{\gamma}_0 \sim \textbf{Unif}\left(\left[0, \eta\right]^{N}, \mathbb{I}_{N\times N}\right)$
\STATE $S(0) \gets \argmax_{S \subset [N]:\atop |S|=C} \sum_{j\in S}  \hat n_{j}(0) + \gamma_{0,j}$
\STATE $\mathrm{flag}\gets 0$
\FOR{\( t = 1 \) to \( T \)}
    \STATE $\beta_t \sim \textbf{Bernoulli}(q) $
    \IF{$\delta_t = 1$ and $\beta_t=1$}
        \STATE $\hat n_{f_t}(t)\gets \hat n_{f_t}(t-1) + 1$
        \STATE $\mathrm{flag}\gets 1$
        \ENDIF
    \IF{$t \% B =0$ and $\mathrm{flag}=1$} \label{line:ConditionBatchSample}
            \STATE $\boldsymbol{\gamma}_t \gets  \textbf{UpdateNoiseVectors}(\boldsymbol{\gamma}_0, \eta, \hatbf{n}(t)) \; (\boldsymbol{\gamma}_t\sim \boldsymbol{\gamma}_0)$ 
            \STATE $S(t) \gets \argmax_{S\subset [N]:\atop |S|=C} \sum_{j\in S}  \hat n_{j}(t) + \gamma_{t,j}$ \label{line:decison-rule}
            \STATE $\mathrm{flag}\gets 0$
        \ELSE
            \STATE $S(t)\gets S(t-1)$
    \ENDIF
\ENDFOR
\end{algorithmic}
\label{algo:NFPL-singleCache}
\end{algorithm}
All these algorithms are derived from the same family that we call \texttt{NFPL}, whose dynamics are described in Algorithm~\ref{algo:NFPL-singleCache}. At step $t=0$, an \texttt{NFPL} algorithm generates a noise vector~$\boldsymbol{\gamma}_0 = (\gamma_{0,f})_{f\in \mathcal{I}}$ from the multivariate uniform distribution with uncorrelated components, constrained within the range $[0, \eta]^N$, where $\eta= \sqrt{BT/2C}$. Algorithms from this family maintain approximate counts of past requests for each file, denoted $\hatbf{n}(t)= (\hat{n}_{f}(t))_{f\in \mathcal{I}}$, and expressed as $\hat{n}_{f}(t) = \sum_{t=1}^{T} \delta_t \beta_t \cdot \mathds{1}(f_t = f)$, where~$\delta_t$ and~$\beta_t$ are Bernoulli random variable with success probabilities denoted~$p$ and~$q$, respectively. The variable $\delta_t$ indicates the observability of the request at step~$t$, and~$\beta_t$ is an algorithm-generated variable to reduce the frequency of counter updates. Moreover, the cached content is only updated when the time step~$t$ is a multiple of~$B$ and if there has been at least one request between the time steps $(k-1)B+1$ and $kB$, where $t=kB$, such that $\delta_t \beta_t = 1$. In these time steps, a noise vector, denoted~$\boldsymbol{\gamma}_t$, with a probability distribution identical to that of~$\boldsymbol{\gamma}_0$, is used to decide cache content. Specifically, the~$C$ files with the highest perturbed approximate counts: $\hat{n}_{f}(t) + \gamma_{t,f}$, for $f \in \mathcal{I}$, are stored in the cache. We consider three variants of \texttt{NFPL} that differ in the temporal correlations between the noises $(\boldsymbol{\gamma}_t)_{t\in [T]}$:

    \begin{enumerate}
        \item \textbf{S-NFPL.} The noise vector remains static over time, i.e. $\boldsymbol{\gamma}_t= \boldsymbol{\gamma}_0, \; \forall t\in [T]$.
        

        \item \textbf{D-NFPL.}  The noise vector is regenerated independently from previously used noise vectors from the multivariate uniform distribution with uncorrelated components, constrained within the range $[0, \eta]^N$, i.e., $(\boldsymbol{\gamma}_t)_{t\in [T]}$ are i.i.d. multivariate uniform random variables.


        \item \textbf{L-NFPL.} The noise vectors are correlated and depend on~$\boldsymbol{\gamma}_0$ and the approximate counts $\hatbf{n}(t)$ as follows, 
            \begin{align}\label{e:Noise-FLL}
                    \boldsymbol{\gamma}_t = \boldsymbol{\gamma}_0 + \eta \ceil*{\frac{\hatbf{n}(t) -\boldsymbol{\gamma}_0}{\eta}} - \hatbf{n}(t). 
            \end{align}
    \end{enumerate}

\subsection{Regret guarantees }
\label{ss:regret_guarantees}

Any \texttt{NFPL} caching policy, with no assumption about the temporal correlations between the noise vectors $(\gamma_{t})_{t=1}^{T}$, enjoys~$\mathcal{O}(\sqrt{T})$ regret, as shown in Theorem~\ref{th:RegretNFPLMain}. 

\begin{theoremSp} \label{th:RegretNFPLMain}
All variants of \texttt{NFPL} achieve the same regret, that is sublinear:
    \begin{align}
         \mathcal{R}_{T}(\texttt{NFPL}) \leq \frac{2\sqrt{2BC}}{pq} \left(\sqrt{T} + \frac{B}{2} \frac{1}{\sqrt{T}}\right).
    \end{align}
\end{theoremSp}
\vspace{2mm}

\noindent \textbf{Difficulty in proving Theorem~\ref{th:RegretNFPLMain}.} When~$pq=1$, the actual request count for each file is available, and one can prove that \texttt{NFPL} is an instance of \texttt{FPL} with uniform noise, applied for a specific online linear learning problem modeling the caching problem, which allows to use the result in~\cite[Theorem 1.1 a)]{kalai2005efficient}. Note that this specific modeling allows us to improve the constants in the regret bound concerning the \texttt{FPL} caching policy in~\cite{bhattacharjee2020fundamental}.

However, when $pq<1$, \texttt{NFPL}'s update rule is different from an \texttt{FPL}-based caching policy due to the use of approximate request counts~$\hatbf{n}$ preventing us from directly using online learning results for \texttt{FPL}. Furthermore, it is not straightforward to relate the probability distribution of the content selected by \texttt{NFPL} ($S_{\texttt{NFPL}}(t)$) to that of \texttt{FPL} ($S_{\texttt{FPL}}(t)$).

This complexity is evident even in the simple case of two files, and a cache capacity \( C = 1 \). Here, the cache can either hold file \(\{1\}\) or file \(\{2\}\). For \texttt{FPL}, the probability of caching file 1 is equal to~\(\Proba{\gamma_{t,1} - \gamma_{t,2} > n_2 - n_1}\), which can be derived in closed form because \(\gamma_{t,1}\) and \(\gamma_{t,2}\) are independent and uniformly distributed. However, in \texttt{NFPL}, the corresponding probability is \(\Proba{\gamma_{t,1} - \gamma_{t,2} - \hat{n}_2(t) + \hat{n}_1(t) > 0}\), which involves a sum of uniform and binomial random variables. This sum does not yield a closed-form distribution, making it challenging to relate this probability to that of \texttt{FPL}, especially for generic values of~$N$ and~$C$.

\begin{IEEEproof}[Sketch of the proof of Theorem~\ref{th:RegretNFPLMain}]
\noindent  The key idea is to consider a variant of the online linear learning framework from Section~\ref{ss:OnlineLearning_FPL}, where, rather than observing the cost vectors~$\mathbf{r}_t$ directly, the agent receives only noisy estimations, denoted as~$\hatbf{r}_t$. This variation accounts for the partial observability inherent in our caching problem. We then demonstrate that under specific assumptions about these estimates, substituting the exact cost vectors with their estimates in the \texttt{FPL} algorithm results in sublinear regret. This approach generalizes the \texttt{NFPL} algorithm to the online linear learning framework. To establish the regret bound for \texttt{NFPL}, we introduce an auxiliary online learning problem, where the update rule for \texttt{FPL} in this auxiliary problem coincides with that of \texttt{NFPL} in the original problem. We leverage this property to show that $\mathcal{R}_T(\texttt{NFPL})\leq \hat{\mathcal{R}}_T(\texttt{FPL})$, where $\hat{\mathcal{R}}_T(\mathcal{A})$ represents the regret of an algorithm~$\mathcal{A}$ in the auxiliary problem. This allows us to prove the sublinear regret property of \texttt{NFPL} both in the general case and, specifically, in the context of caching.

  The detailed proof is presented in the supplementary material (Section~\ref{Proof:Regret-NFPL}).  
\end{IEEEproof}

\vspace{2mm}


\noindent \textbf{NFPL in the general framework of online linear learning.} 
A fundamental step in the proof of Theorem~\ref{th:RegretNFPLMain}
is to show first that, for any online linear learning problem (with caching as a particular case), substituting the exact cost vectors by their respective noisy estimates within the \texttt{FPL} algorithm (a generalization of \texttt{NFPL}) leads to sublinear regret under specific assumptions on the estimates. A simple and notable case of these assumptions holds when $\hatbf{r}_t = b_t \cdot \mathbf{r}_t$, with~$(b_t)_{t}$ being i.i.d. Bernoulli random variables. In this particular case, \texttt{NFPL} can be viewed as a sublinear regret algorithm that limits how often the adversary’s feedback is revealed. This feature of \texttt{NFPL} in this setting makes it a \textit{selective sampling} algorithm~\cite{adaptiveSelectiveExperts24}, also referred to as a \textit{label-efficient forecaster}~\cite{expertWithPartialTruth05}. Such algorithms are motivated by practical scenarios where it is costly to observe the cost vector chosen by the adversary, prompting the need to reduce these observations while maintaining regret guarantees. However, to the best of our knowledge, most label-efficient algorithms in the literature are variations of the multiplicative weights algorithm, typically applied to the expert problem, which is a particular case of the online linear learning framework covered by \texttt{NFPL}.

\vspace{2mm}

 \noindent \textbf{Sublinear regret under partial observation of the requests.} Surprisingly, despite only observing a fraction~$p$ of the past requests, Theorem~\ref{th:RegretNFPLMain} shows that \texttt{NFPL} achieves the asymptotically optimal regret bound~$\mathcal{O}(\sqrt{T})$. Moreover, the factor~$\frac{1}{p}$ in the regret bounds of Theorem~\ref{th:RegretNFPLMain} captures the performance loss of the algorithms due to the partial observation of the requests.

  Even when all past requests are observed, i.e., $p=1$, \texttt{NFPL} has several advantages with respect to the \texttt{FPL}-based caching policy in~\cite{bhattacharjee2020fundamental}.

 \vspace{2mm}
 
 \noindent \textbf{Tightness of the regret bound.} The comparison between~\cite[Thm. 3]{bhattacharjee2020fundamental} and Theorem~\ref{th:RegretNFPLMain} reveals that, by sampling the noise~$\boldsymbol{\gamma}$ from the uniform distribution instead of the Gaussian distribution, our \texttt{FPL}-based caching policies have a regret bound that does not depend on the catalog size~$N$. Consequently, and unlike the \texttt{FPL} caching policy with Gaussian noise, 
 \texttt{NFPL} regret guarantees do not deteriorate when~$N$ goes to infinity. 
Moreover, the lower bound on the regret of any caching policy, established in~\cite[Thm.~2]{bhattacharjee2020fundamental}, is~$\mathcal{O}(\sqrt{CT})$ when~$B=1$, highlighting that our uniform based \texttt{FPL} caching policies enjoy asymptotically optimal regret in~$C$,~$N$ and~$T$.

\vspace{2mm}

 \noindent \textbf{Advantages of the parameters $B$ and $q$.} The parameter~$q$ enables memory reduction, since \texttt{NFPL} only increments the counters~$\hatbf{n}$ at step~$t$ when the Bernoulli random variable~$\beta_t$ is equal to~$1$. As a result, the maximum entry in the absolute value of the vector~$\hatbf{n}(T)+ \boldsymbol{\gamma}_T$ is~$\mathcal{O}(qT)$, in expectation, compared to~$\mathcal{O}(T)$ in the standard \texttt{FPL}. Therefore, \texttt{NFPL} saves memory by storing a vector with smaller entries. Finally, the parameters~$B$ and~$q$ allow to reduce the frequency of cache updates, which can lead to reduced communication cost between the main memory and the cache.

\subsection{Amortized time complexity}
\label{ss:time_complexity}

The three variants of \texttt{NFPL} are characterized by distinct joint distributions of the noise vectors~$(\boldsymbol{\gamma}_t)_{t\in [T]}$. Theorem~\ref{th:TimeComplexityNFPL} shows for each variant the effect of the parameters~$B$ and~$q$ on the amortized time complexity.

\begin{theoremSp}\label{th:TimeComplexityNFPL}
        The amortized time complexity of the three variants of \texttt{NFPL} is as follows:
            \begin{itemize}
                \item \texttt{S-NFPL}: $\mathcal{O}\left( 1+ pq \ln{C} \right)$. 

                \item \texttt{D-NFPL}: $\mathcal{O}\Big(1 + N\ln{N}/B\Big)$.

                \item \texttt{L-NFPL}: $\mathcal{O}\left(1+pq\ln(C)\sqrt{C}/\sqrt{BT}\right)$. 
            \end{itemize}
\end{theoremSp}

\vspace{3mm}


\begin{IEEEproof}[Sketch of the proof]\noindent The most computationally intensive task in \texttt{NFPL} variants is identifying the top~$C$ files based on their perturbed counts, $\hatbf{n} + \boldsymbol{\gamma}$, denoted as~$\mathbf{m}$. In \texttt{D-NFPL}, the noise vectors~$(\boldsymbol{\gamma}_t)_{t \in [T]}$ are consistently regenerated, necessitating a sorting operation to determine the top~$C$ files, which accounts for the~$\mathcal{O}(N \ln{N})$ term in its amortized time complexity. Conversely, in \texttt{S-NFPL} and \texttt{L-NFPL}, only one component of the vector~$\mathbf{m}(t) = \hatbf{n}(t) + \boldsymbol{\gamma}_t$ is updated between consecutive time steps, allowing the use of a \textit{heap} data structure to track the top~$C$ files efficiently. The time complexity of an insertion/deletion in a heap is~$\mathcal{O}(\ln{C})$, explaining the~$\ln{C}$ term in their amortized time complexity. In \texttt{L-NFPL}, it is even possible for $\mathbf{m}(t) = \mathbf{m}(t+1)$, and we show that the probability of this event occurring is on the order of~$\mathcal{O}(\sqrt{C/(BT)})$, which justifies the amortized time complexity of \texttt{L-NFPL}. 

    The detailed proof is presented in the supplementary material (Section~\ref{Proof:TimeComplexityNFPL}).
\end{IEEEproof}
Theorems~\ref{th:RegretNFPLMain} and~\ref{th:TimeComplexityNFPL} show that \texttt{L-NFPL} is the first~$\mathcal{O}(\sqrt{T})$-regret caching policy with~$\mathcal{O}(1)$ amortized time complexity as~$T$ goes to infinity, considerably improving over the best previously achieved amortized time complexity of~$\mathcal{O}(\ln{C})$.

\subsection{Trade-off between the regret and the time complexity}
\label{ss:Tradeoff_regret_time}

We discuss in this section how batching, i.e., updating the cache after~$B$ requests, and sampling the requests with rate~$q$, balance the regret. Finally, we conclude by comparing the three \texttt{NFPL} variants in terms of the variance of the miss ratio, which is not reflected in the regret metric expressed in~\eqref{e:defRegretCaching}, where only the expectation is considered.

\vspace{2mm}

\noindent \textbf{Batching.} Previous works proposed batching to reduce the computational cost of no-regret caching policies by roughly a factor of~$1/B$ and resulting in a regret bound of~$\mathcal{O}(\sqrt{BT})$~\cite{si2023no,MhaisenSigmetrics}. Our paper is the first to extend a similar result for an \texttt{FPL}-based caching policy when the noise vector is regenerated in an i.i.d. fashion (\texttt{D-NFPL}), leading to a regret bound of $\mathcal{O}(\sqrt{BT})$ and an amortized time complexity of~$\mathcal{O}(N\ln{N}/B)$. However, batching has a marginal effect on the amortized time complexity of \texttt{S-NFPL}.

\vspace{2mm}

\noindent \textbf{Sampling.} Omitting a fraction~$q$ of past requests from the decision-making process in \texttt{S-NFPL} allows the amortized time complexity to be lowered to~$\mathcal{O}(1+ q\ln{C})$, instead of~$\mathcal{O}(\ln{C})$, but results in a regret bound that grows by a factor of~$1/q$. A similar trade-off occurs in \texttt{L-NFPL}, where the regret bound is~$\mathcal{O}(\frac{1}{q}\sqrt{T})$, and the amortized time complexity is~$\mathcal{O}(1 + \frac{q}{\sqrt{T}})$, ignoring the dependence on~$C$ and~$B$. Consequently, for a no-regret caching policy that employs batching to match \texttt{L-NFPL}'s computational cost, $B$ would need to be set to~$\mathcal{O}(\frac{\sqrt{T}}{q})$, leadind to a regret bound of~$\mathcal{O}(\sqrt{\frac{1}{q}} T^{3/4})$, significantly worse than \texttt{L-NFPL}'s~$\mathcal{O}(\frac{1}{q}\sqrt{T})$ regret for large values of~$T$. 



\vspace{2mm}
\noindent \textbf{Variance.} Although our theoretical results indicate that \texttt{L-NFPL} is the superior choice among the variants, our numerical simulations reveal that \texttt{D-NFPL} exhibits lower variability around the average cost, suggesting that it is more likely to perform consistently close to the average. To understand this aspect, we stress that \texttt{FPL}-based caching policies differ from \texttt{LFU} by  randomly assigning a score~($\boldsymbol{\gamma}$) to files that reflect their importance, independently of the request count, and then combining this score with the request count to decide if the file is stored. This allows the policy to store not only frequently requested files but also to explore other potentially relevant files, such as those that experience bursts of requests within a specific window of time but do not consistently rank among the top~$C$ files. In \texttt{S-NFPL} and \texttt{L-NFPL}, the score~($\boldsymbol{\gamma}$) is determined once and remains fixed, whereas in \texttt{D-NFPL}, the score is continuously refined, enabling broader exploration. For instance, files initially given a high score in~$\boldsymbol{\gamma}$ might later prove irrelevant, causing a drop in the performance of the \texttt{FPL} caching policy in comparison to \texttt{LFU}. However, by consistently updating the scores, \texttt{D-NFPL} ensures that relevant files are more likely to be selected, even if they are occasionally missed.

%% file: Journal/experiments.tex
\begin{figure*}[t]
\centering

\begin{subfigure}{0.32\linewidth}
  \centering
  \includegraphics[width=0.99\linewidth, keepaspectratio]{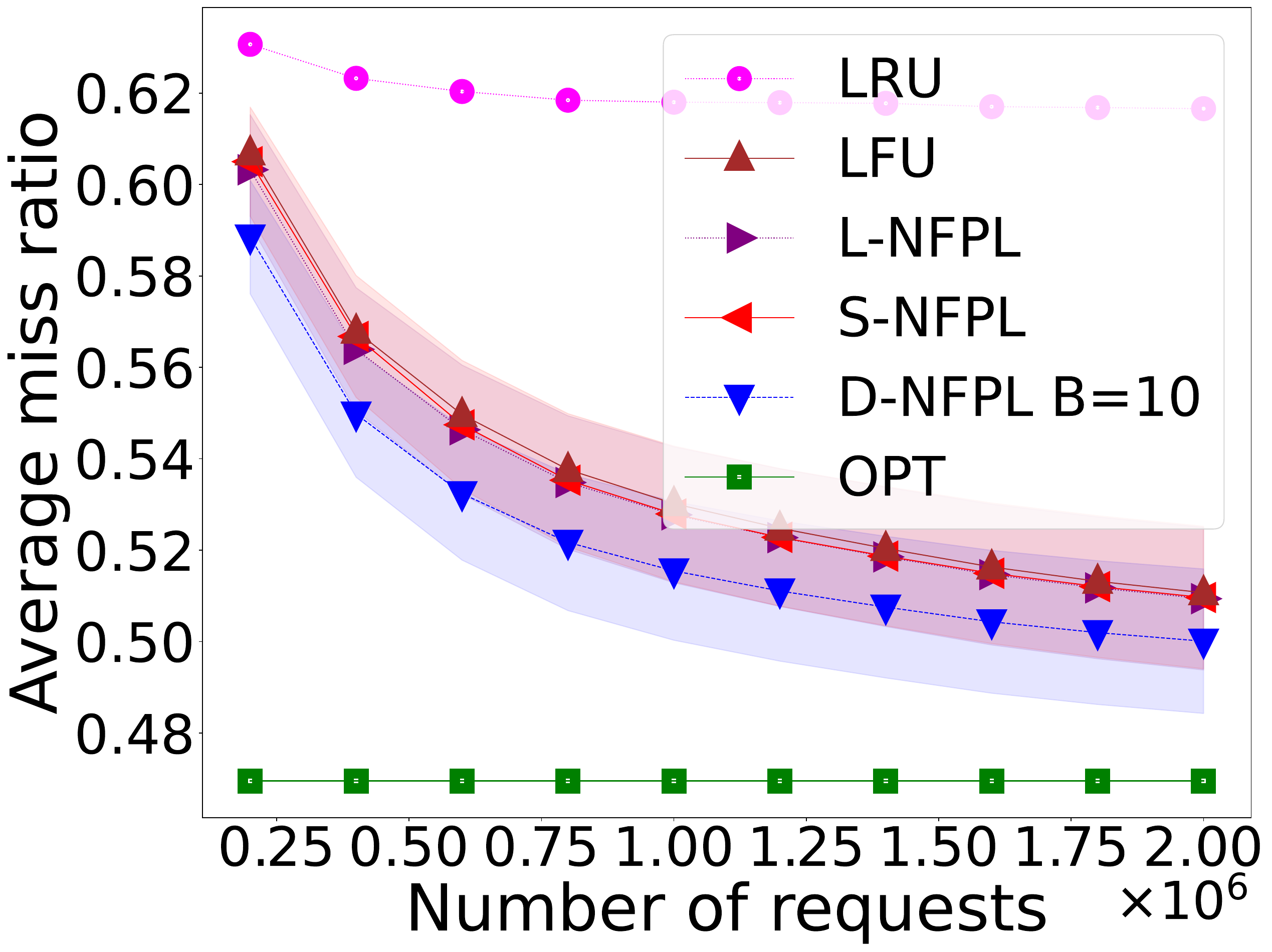}
  \caption{Zipf ($p = 0.01$)}
  \label{fig:zipf-p0.01}
\end{subfigure}
\hfill
\begin{subfigure}{0.32\linewidth}
  \centering
  \includegraphics[width=0.99\linewidth, keepaspectratio]{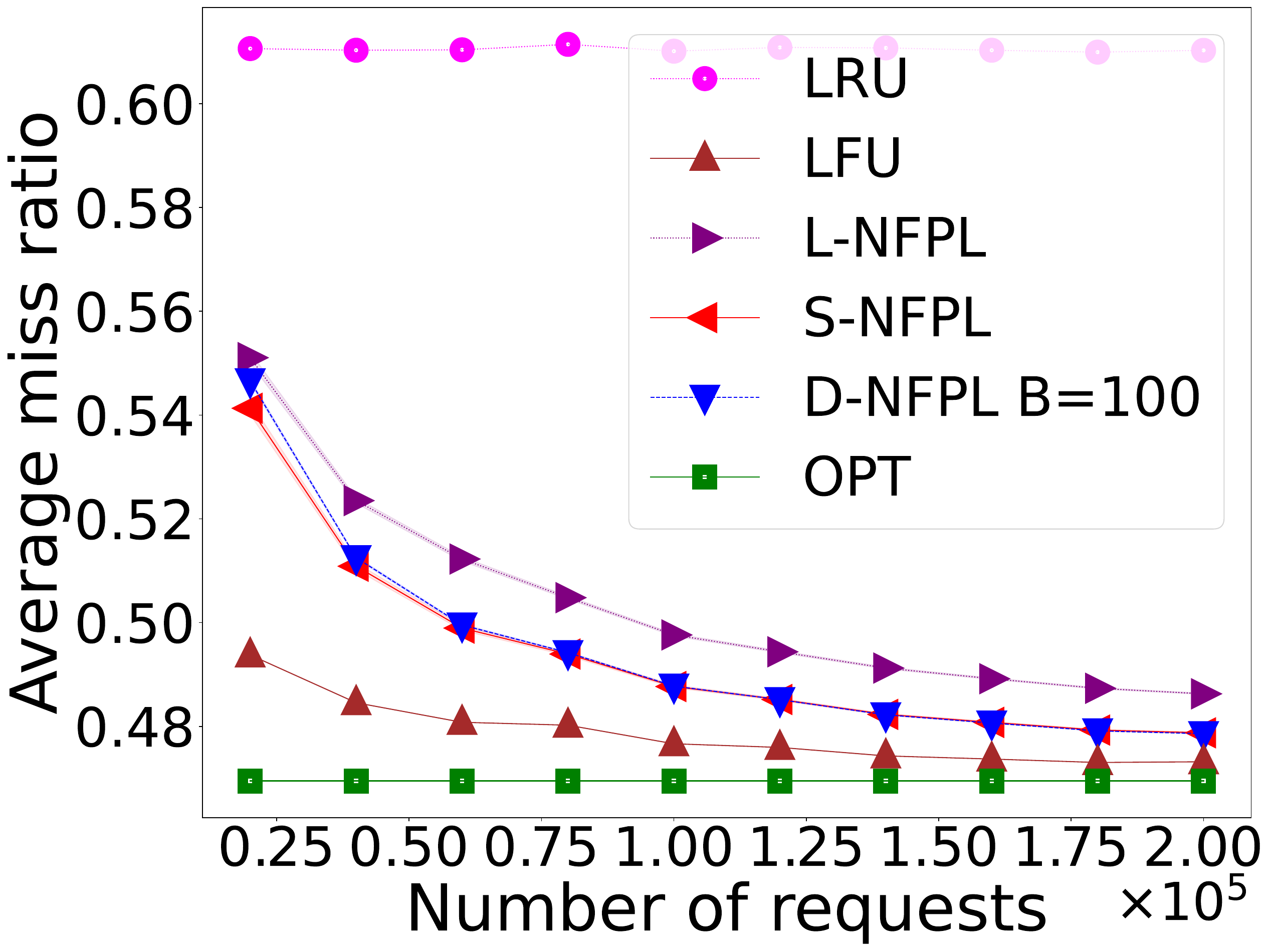}
  \caption{Zipf ($p = 1)$}
  \label{fig:zipf-p1}
\end{subfigure}
\hfill
\begin{subfigure}{0.32\linewidth}
  \centering
  \includegraphics[width=0.99\linewidth, keepaspectratio]{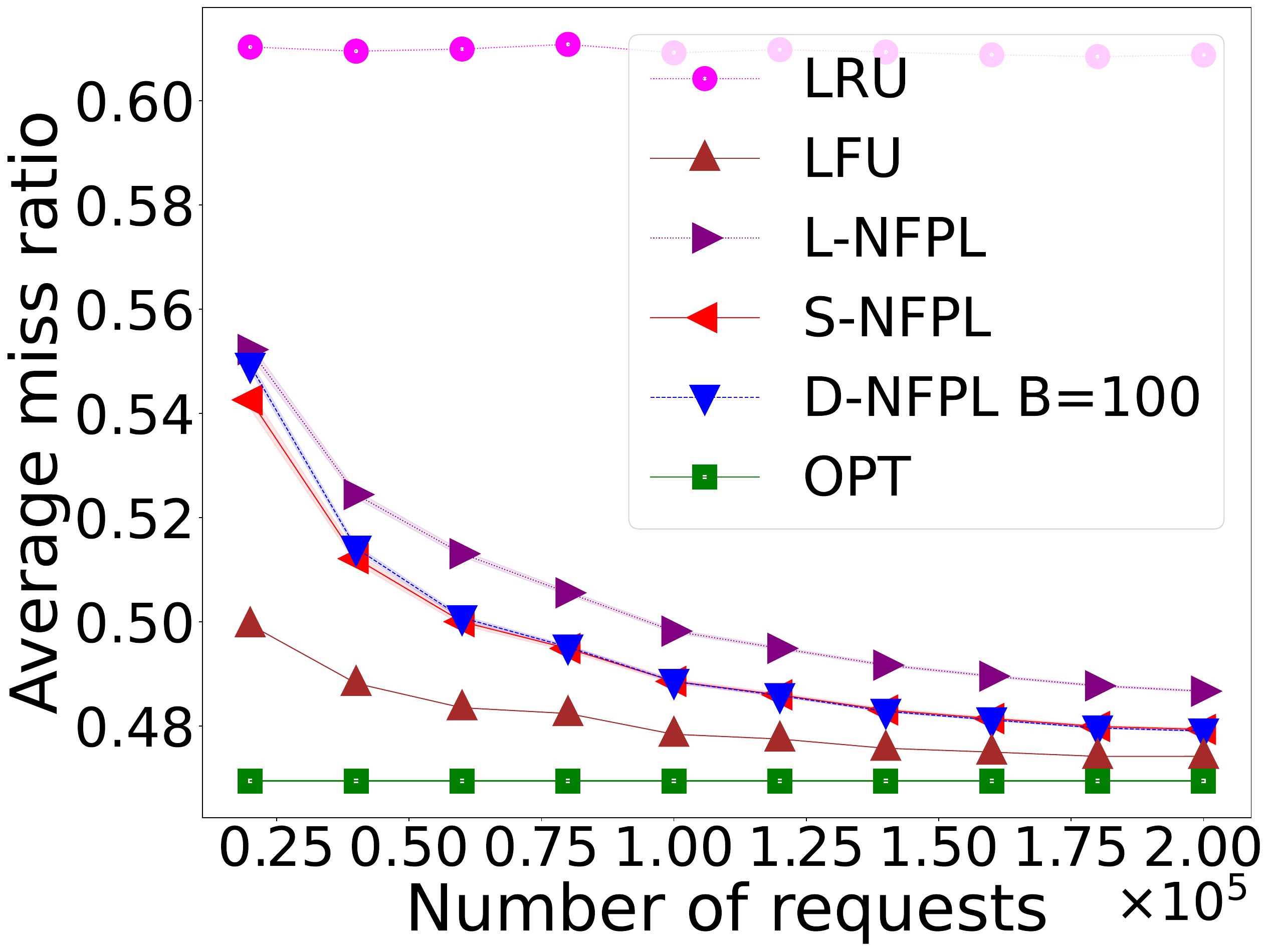}
  \caption{Zipf ($p = 0.7$)}
  \label{fig:zipf-p0.5}
\end{subfigure}

\vspace{0.3cm}

\begin{subfigure}{0.32\linewidth}
  \centering
  \includegraphics[width=0.99\linewidth, keepaspectratio]{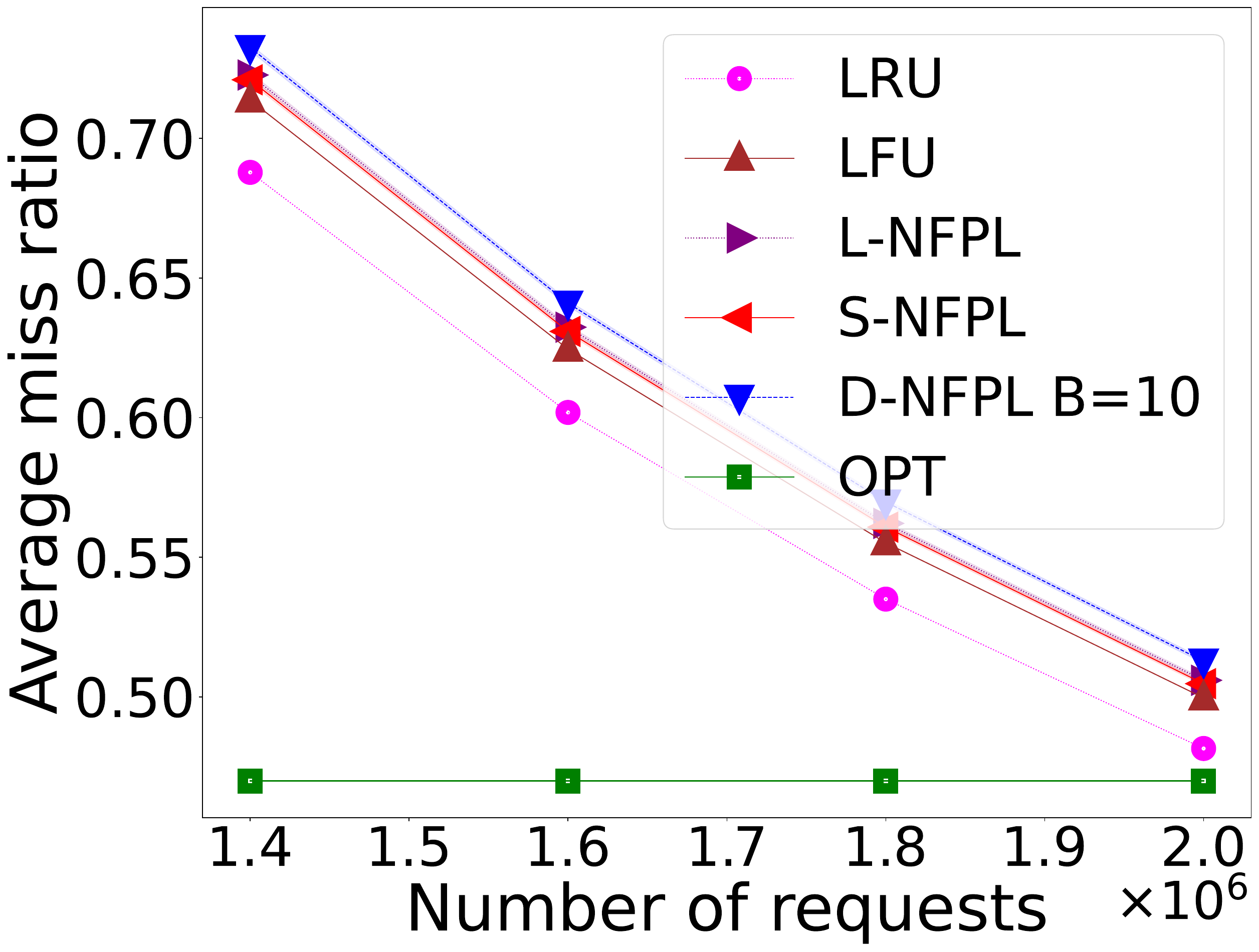}
  \caption{Zipf-RR ($p = 0.01$)}
  \label{fig:zipfrr-p0.01}
\end{subfigure}
\hfill
\begin{subfigure}{0.32\linewidth}
  \centering
  \includegraphics[width=0.99\linewidth, keepaspectratio]{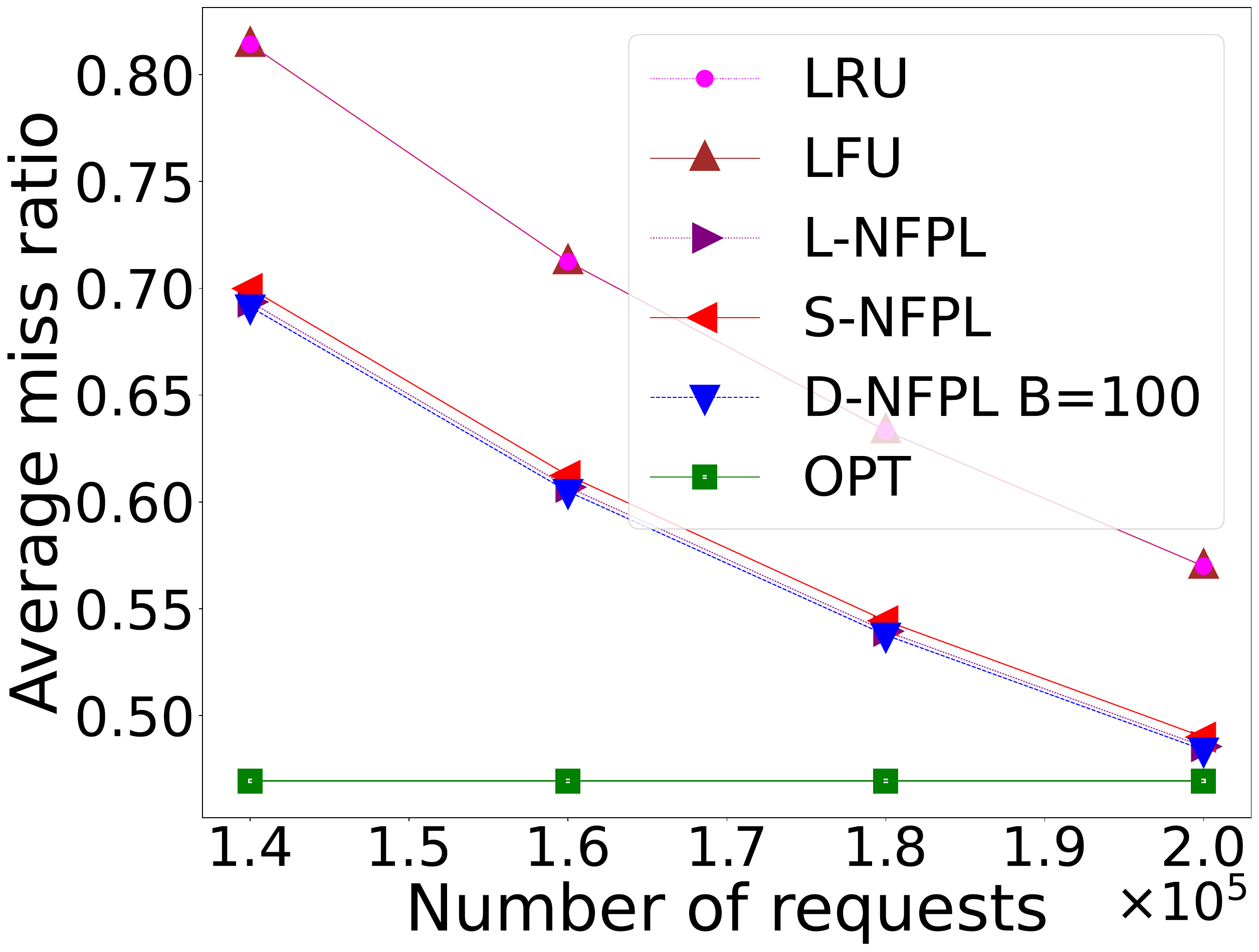}
  \caption{Zipf-RR ($p = 1$)}
  \label{fig:zipfrr-p1}
\end{subfigure}
\hfill
\begin{subfigure}{0.32\linewidth}
  \centering
  \includegraphics[width=0.99\linewidth, keepaspectratio]{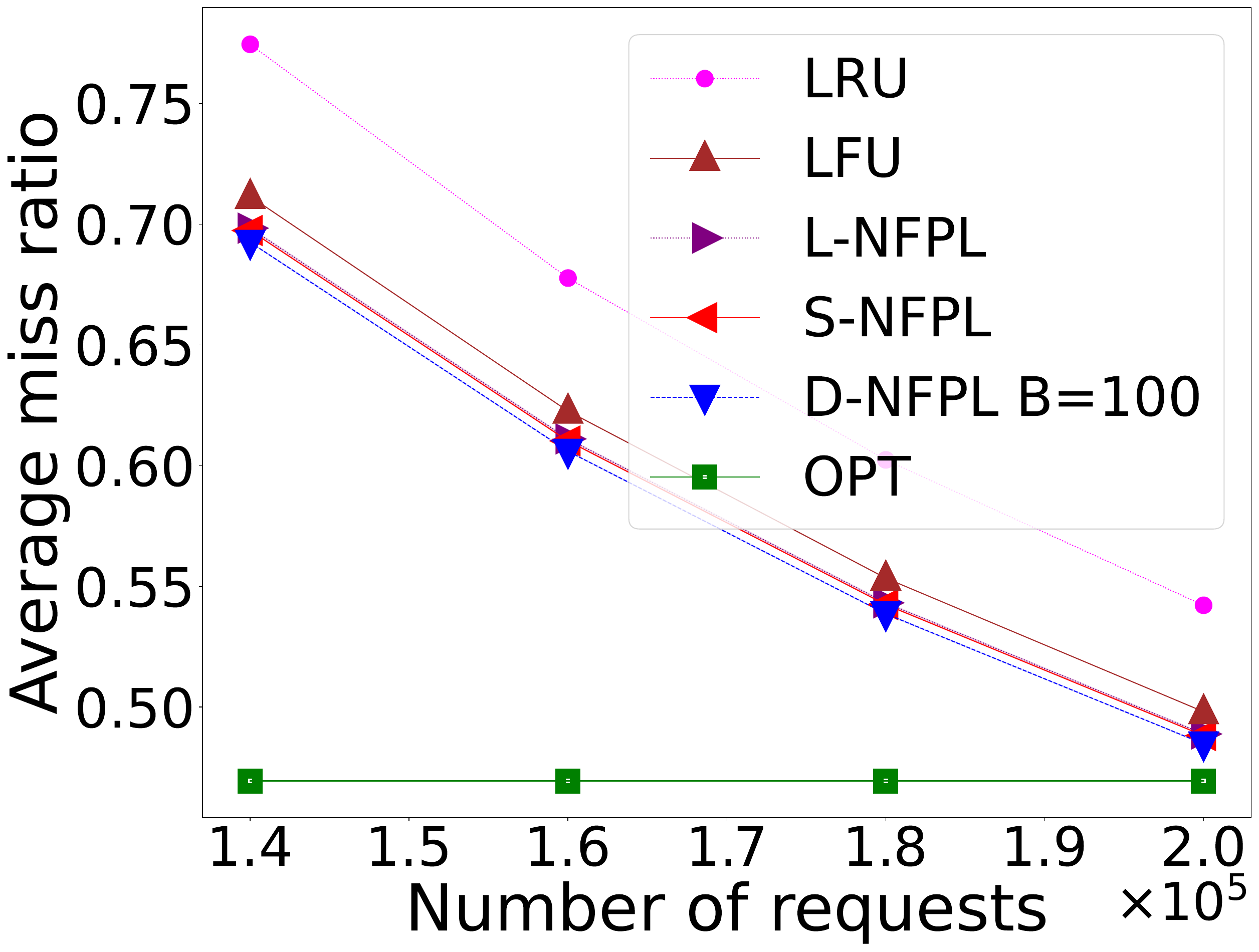}
  \caption{Zipf-RR ($p = 0.7$)}
  \label{fig:zipfrr-p0.7}
\end{subfigure}

\vspace{0.3cm}

\begin{subfigure}{0.32\linewidth}
  \centering
  \includegraphics[width=0.99\linewidth, keepaspectratio]{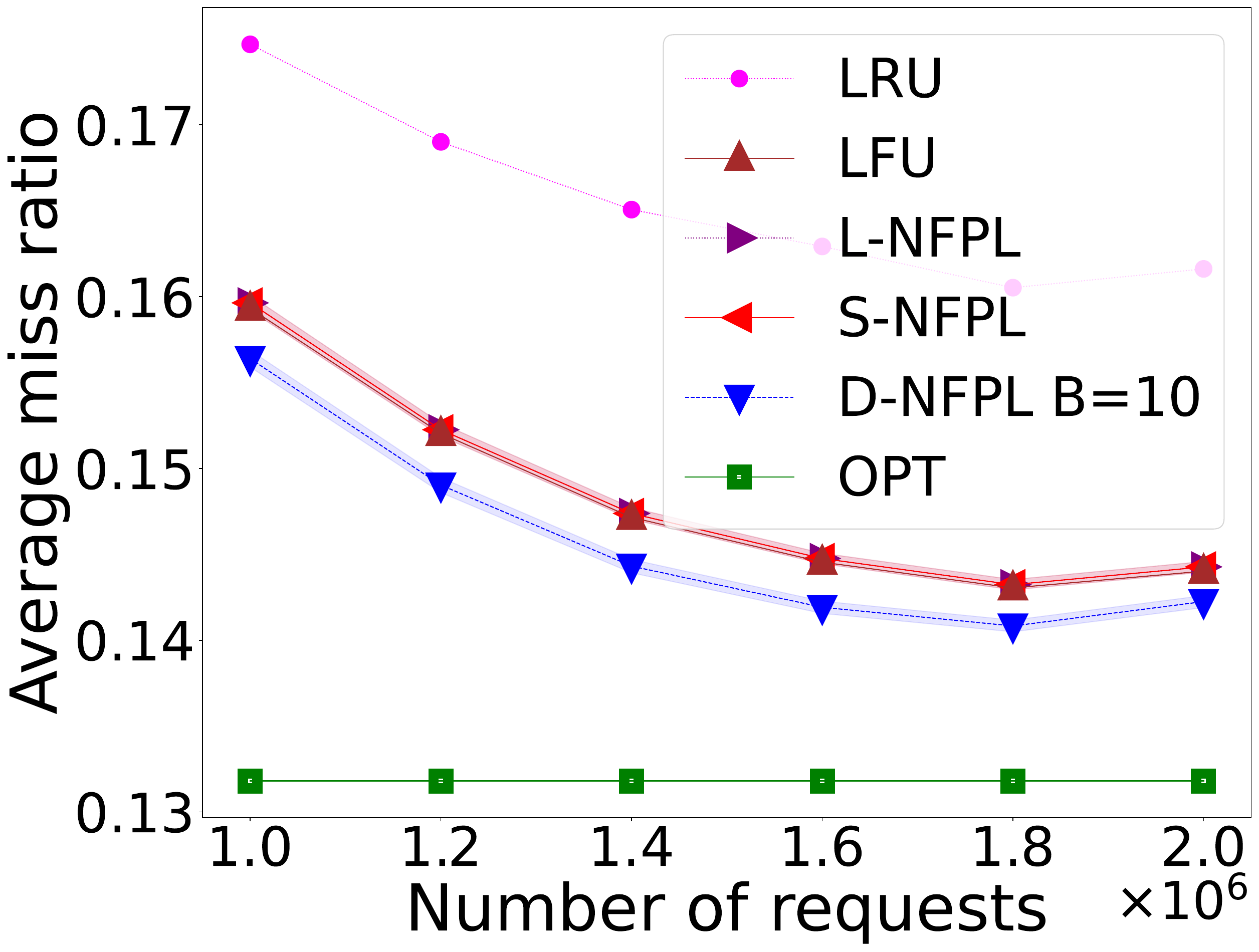}
  \caption{Akamai ($p = 0.01$)}
  \label{fig:akamai-p0.01}
\end{subfigure}
\hfill
\begin{subfigure}{0.32\linewidth}
  \centering
  \includegraphics[width=0.99\linewidth, keepaspectratio]{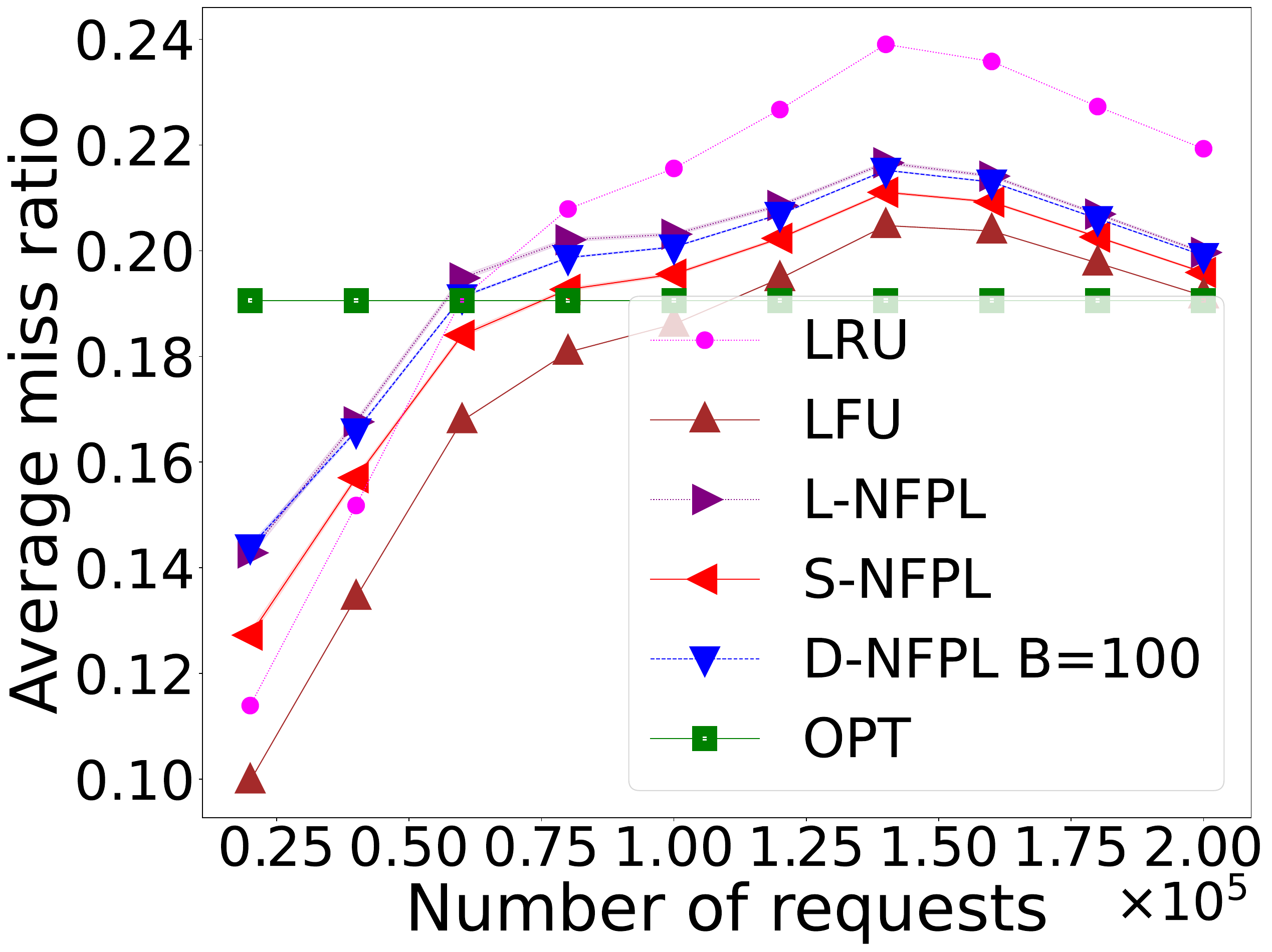}
  \caption{Akamai ($p = 1$)}
  \label{fig:akamai-p1}
\end{subfigure}
\hfill
\begin{subfigure}{0.32\linewidth}
  \centering
  \includegraphics[width=0.99\linewidth, keepaspectratio]{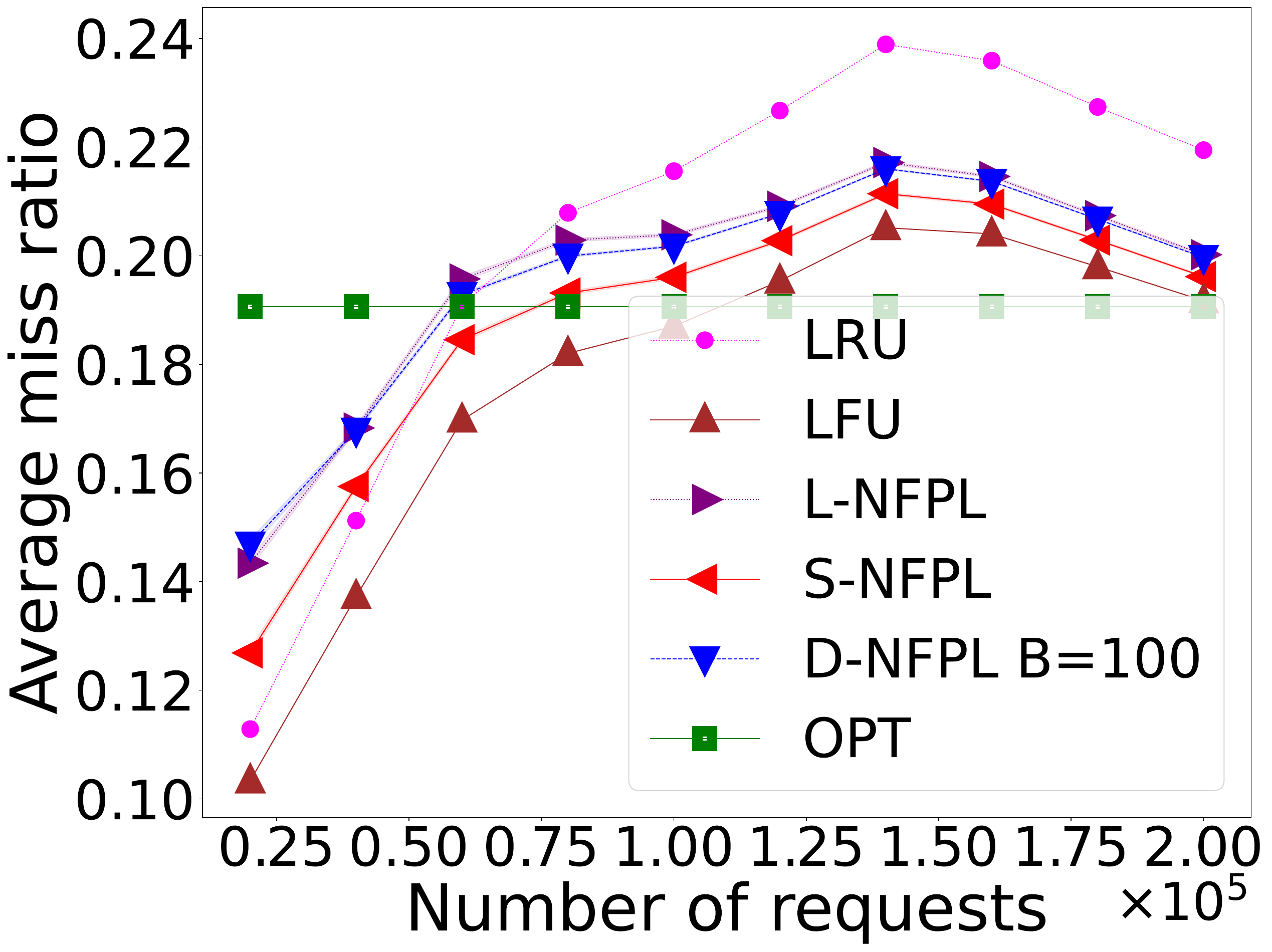}
  \caption{Akamai ($p = 0.7$)}
  \label{fig:akamai-p0.7}
\end{subfigure}

\caption{Average miss ratio for different values of $p$ across caching policies and traces.}
\label{fig:miss-ratio}
\vspace{-0.1in}
\end{figure*}

\begin{figure*}[t]
\centering
\begin{subfigure}{0.32\linewidth}
    \centering
    \includegraphics[width=0.95\linewidth, keepaspectratio]{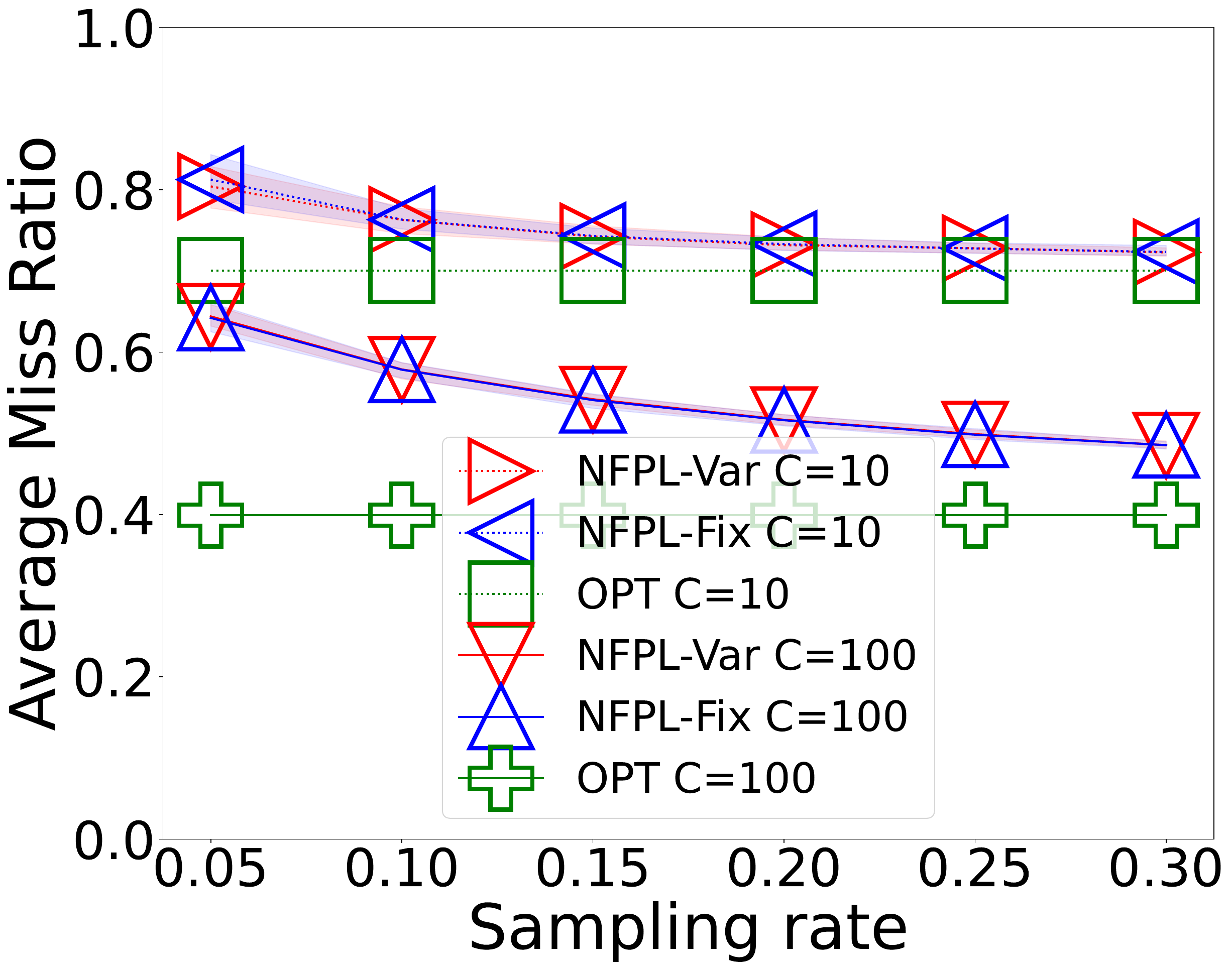}
\caption{Zipf}
\label{fig:Zipf-NFPL-FIX-VAR}
\end{subfigure}
\hfill
\begin{subfigure}{0.32\linewidth}
  \centering
  \includegraphics[width=0.95\linewidth,keepaspectratio]{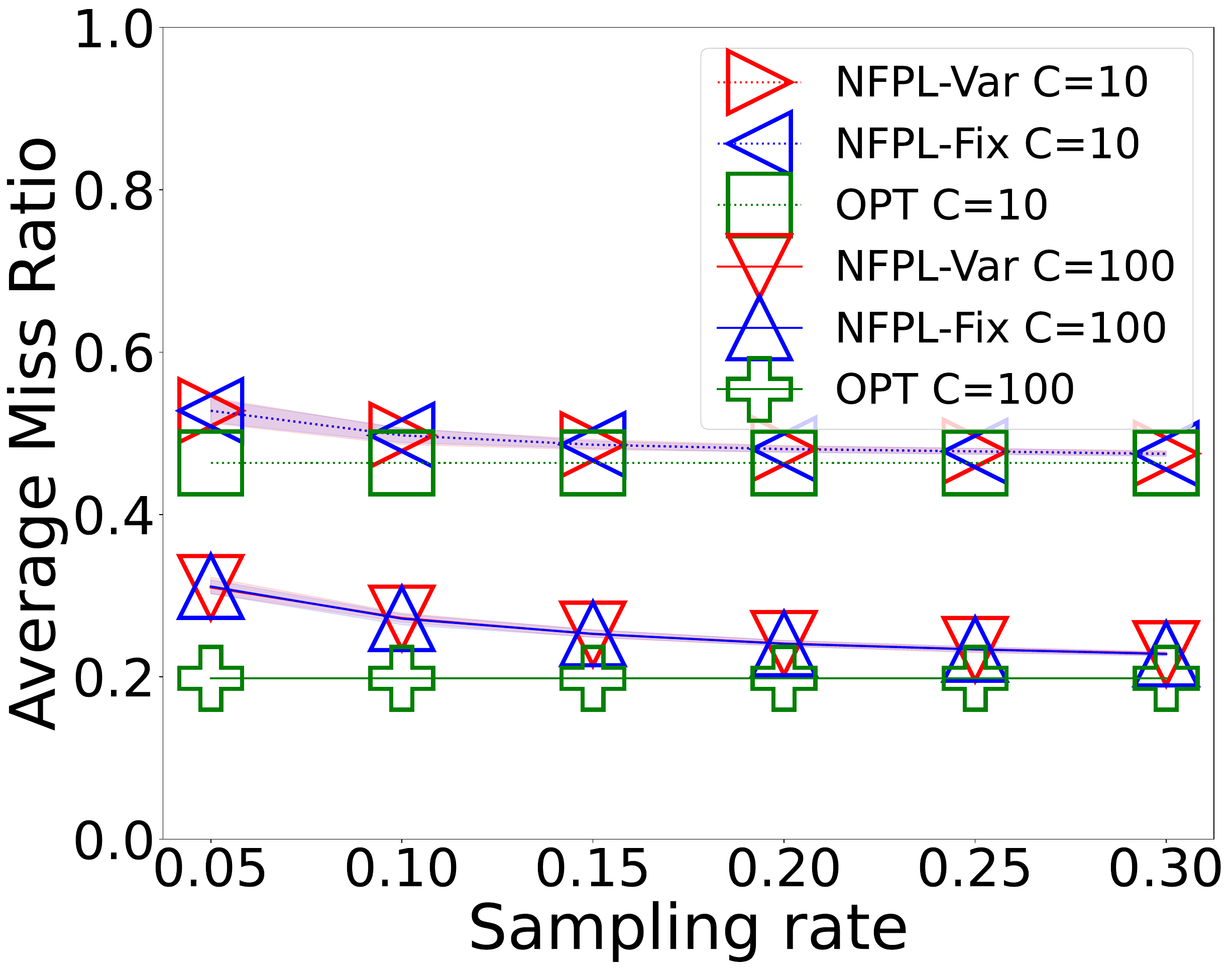}
\caption{Akamai}
\label{fig:Akamai-NFPL-FIX-VAR}
\end{subfigure}
\hfill
\begin{subfigure}{0.32\linewidth}
  \centering
  \includegraphics[width=0.95\linewidth,keepaspectratio]{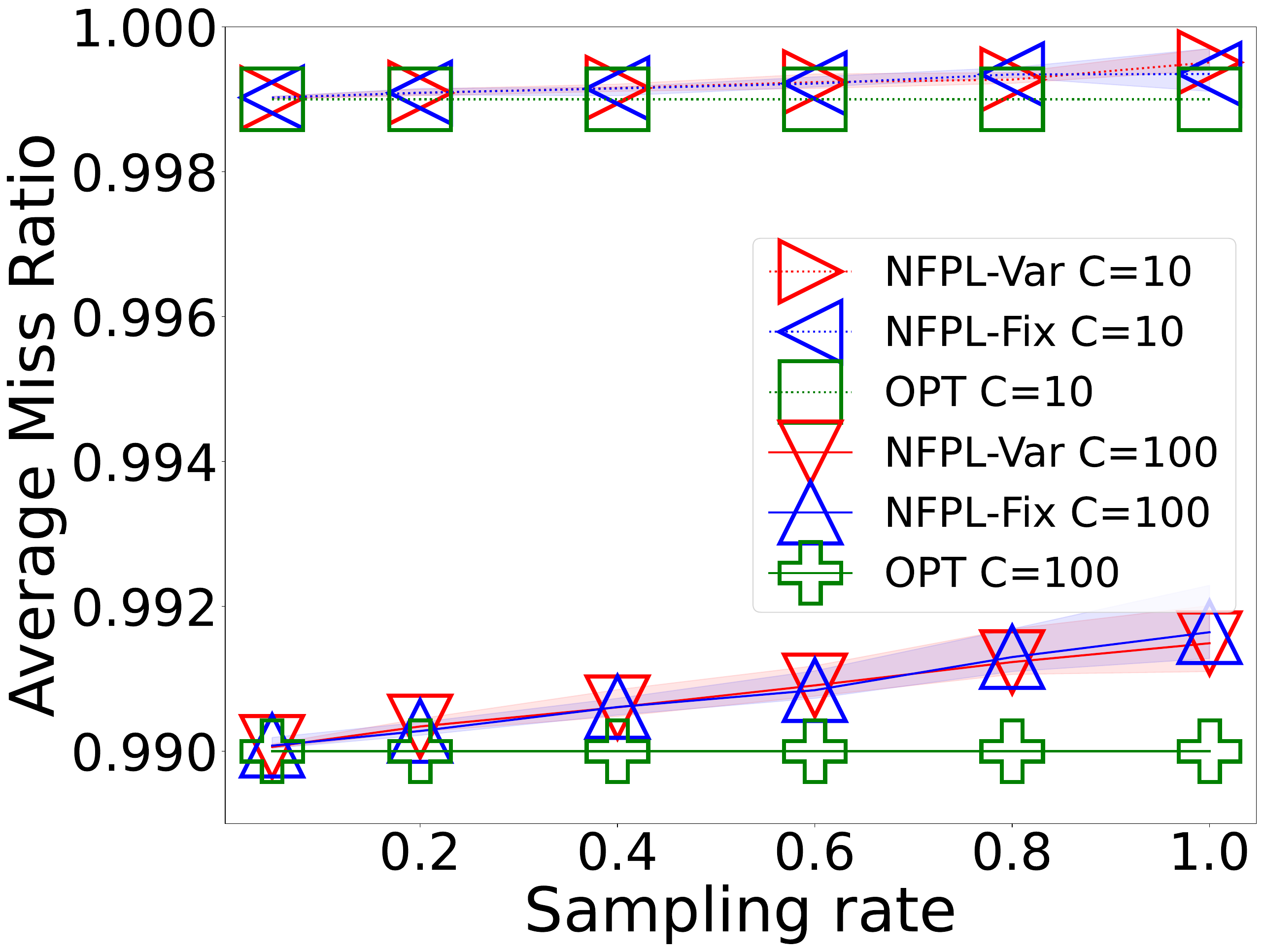}
\caption{Round-robin}
\label{fig:Round-robin-NFPL-FIX-VAR}
\end{subfigure}
\caption{Average miss ratio vs. sampling probability.}
\label{fig:NFPL-FIX-VAR}
\vspace{-0.1in}
\end{figure*}
\color{black}
We conducted simulations of the \texttt{NFPL} algorithms and other existing policies, using both synthetic and real-world traces. Details about the traces are presented in Section~\ref{ss:Datasets}, while Section~\ref{ss:Caching-policies} discusses the caching baselines. In Section~\ref{ss:NFPL-VS-ClassicalPolicies}, we evaluate the effectiveness of our proposed algorithms, in terms of the average miss ratio, the variance around this quantity, and the computational cost. Finally, we show in Section~\ref{ss:NFPL-Fix-Vs-NFPL-Var} the effect of sampling on the performance of the \texttt{NFPL} algorithm.



\subsection{Traces}
\label{ss:Datasets}

{\noindent  \bf{Zipf trace.}} We generate the requests from a catalog of~$N=10^{4}$ files, with $I=[N]$, following an i.i.d. Zipf distribution with exponent~$\alpha=1$, i.e., $(f_t)_{t=1}^{T}$ are i.i.d. random variables and $\Proba{f_t= i}\sim 1/i$. The Zipf distribution is a popular model for the request process in caching~\cite{breslau1999web}.  

\vspace{2mm}

{\noindent  \bf{Zipf-RR trace.}} In this scenario, the catalog size is $N=10^{4}$. The total number of requests for each file, $(n_{i}(T))_{i\in \mathcal{I}}$, follows the multinomial distribution, with the number of trials being $T$, and the probability associated with each file is given by the Zipf distribution with exponent $\alpha=1.0$. Next, the files are numbered based on their total number of requests, with file~$1$ having the most requests and file~$N$ the fewest. The order in which these requests occur follows a specific pattern. Initially, the requests begin with file~$N$ and proceed in descending order to file~$1$. This cycle, from $N$ down to $1$, repeats until all requests for file $N$ are exhausted. Once file $N$ has no more requests, the sequence shifts to files~$N-1$ down to~$1$, repeating this cycle until the requests for file~$N-1$ are depleted. This process continues in the same manner for the remaining files. Eventually, when only file~$1$ remains, repeated requests for file~$1$ are made until all of its requests are depleted.


\color{black}

\vspace{2mm}

{ \noindent \bf{Akamai trace.}} 
The request trace, sourced from Akamai CDN as documented in \cite{neglia2017access}, encompasses several days of file requests, amounting to a total of $2\times 10^{7}$ requests for a catalog comprising~$N=10^{3}$ files.


\subsection{Caching policies}
\label{ss:Caching-policies}

We compare our \texttt{NFPL} policies, with the optimal static cache allocation with hindsight (\texttt{OPT}), and two classic caching policies: Least-Frequently-Used (\texttt{LFU}) and Least-Recently-Used (\texttt{LRU}). Upon a miss, \texttt{LFU} and \texttt{LRU} evict from the cache the least popular file and the least recently requested file, respectively. All these caching policies operate under the assumption of the partial observability regime, where each request is observed with a probability~$p$. Moreover, the performance metric is the \textit{average miss ratio} computed as follows
\begin{align}
 \frac{1}{T} \sum_{t=1}^{T} \mathds{1}(f_t \notin \mathcal{S}_{\mathcal{A}}(t)).
\end{align}
The average miss ratio is averaged over~$M=50$ runs, considering different samples of the partial observability regime and the noise vectors $\{\boldsymbol{\gamma}_t\}_{1}^{T}$ in the \texttt{NFPL} policies. To account for the variability across the runs, we report the~$95\%$ confidence intervals. 



\subsection{NFPL vs. classical policies}
\label{ss:NFPL-VS-ClassicalPolicies}

In this section, we present the results for the previously discussed traces and policies, focusing on how the average miss ratio evolves as the number of requests increases. The parameter~$p$ of the BPO regime takes values in $\{ 0.01, 0.7, 1.0 \}$. The cache capacity $C$ is equal to~$100$ for all the caching policies. Moreover, the total number of requests considered in each trace is $T=2\times 10^{5}$ when $p\in \{0.7, 1\}$ and $T=2\times 10^{6}$ when~$p=0.01$. The \texttt{NFPL} policies are configured with~$q=1$. The parameter~$B$ of \texttt{L-NFPL} and \texttt{S-NFPL} is set to~$1$, whereas its value in \texttt{D-NFPL} is $B=10$ when $p=0.01$ and $B=100$ when $p\in  \{0.7, 1\}$. The parameter $\eta$ is set to $p\sqrt{BT/2C}$ for all the \texttt{NFPL} variants. The results are presented in Figure~\ref{fig:miss-ratio}.

\vspace{3mm}
\noindent \textbf{Zipf.} \texttt{D-NFPL} outperforms all the other policies when the sampling probability is low ($p=0.01$). In this scenario, it takes a long time for the request count of each item to become a reliable predictor of future demands, which makes \texttt{LFU} less effective. Additionally, while \texttt{L-NFPL} and \texttt{S-NFPL} do not rely solely on the observed popularity of files, they use a single realization of the noise vector~$\boldsymbol{\gamma}$ to explore other potentially relevant files. This results in a lower rate of exploration compared to \texttt{D-NFPL}, where the noise vector is continually regenerated. In the other regimes where $p=0.7$ and $p=1.0$, \texttt{LFU} rapidly discerns the most popular files and subsequently converges to \texttt{OPT}, because files popularity does not change over time. However, due to the noise $\boldsymbol{\gamma}_t$, the \texttt{NFPL} policies require a longer duration to determine the files to be stored accurately, especially \texttt{L-NFPL}, because of its lazy update rule. Nevertheless, all the \texttt{NFPL} policies outperform \texttt{LRU}, whose miss ratio fails to converge to \texttt{OPT}.

\vspace{3mm}

\noindent \textbf{Zipf-RR.} When $p \in \{0.7, 1.0\}$, the \texttt{NFPL} variants achieve a lower average miss ratio than \texttt{LRU} and \texttt{LFU}. This advantage is particularly noticeable when~$p=1$, where all the \texttt{NFPL} variants yield a similar average miss ratio of~$0.48$, while \texttt{LRU} and \texttt{LFU} also produce similar results of approximately~$0.57$. Hence, the \texttt{NFPL} variants achieve a relative improvement of roughly~$16\%$ over the \texttt{LRU}/\texttt{LFU} group. To justify the superiority of the \texttt{NFPL} policies in this scenario, we divide the trace into two segments. The first segment, where the file ranked~$C+1$ is still being requested, and the second one, where requests for this file stop. During the first segment, the specific order of the requests is such that the next request is neither of the~$C$ most popular files so far nor the~$C$ recently requested items so far, resulting in most of the requests being missed by both \texttt{LFU} and \texttt{LRU}.
In contrast, in the second segment, \texttt{LFU} and \texttt{LRU} achieve perfect performance, with every request being a hit since only the top~$C$ files are requested. Despite the ideal performance in the second segment, the large number of misses for the popular files in the first segment prevents \texttt{LRU} and \texttt{LFU} from converging to the optimal solution. In contrast, the \texttt{NFPL} variants can handle this challenging trace and successfully converge to the optimum, as predicted by our theoretical results. On the other hand, when~$p=0.01$, the adversarial aspect of the observed trace fades, positioning \texttt{LRU} as the leading policy, outperforming all others. Nonetheless, the \texttt{NFPL} policies are guaranteed to reach the miss ratio of \texttt{OPT} for a large enough number of requests.

\color{black}

\vspace{3mm}

\noindent \textbf{Akamai:} In the real-world trace, when~$p=0.01$, the \texttt{D-NFPL} policy outperforms all other caching policies. Indeed, as discussed in the Zipf trace, when the sampling probability is low, the request count becomes an unreliable predictor of future requests rendering \texttt{LFU} inefficient. Meanwhile, the significant loss of information damages \texttt{LRU} as well. In contrast, for the regimes where~$p \in \{0.7, 1\}$, all policies demonstrate comparable performance, with \texttt{LFU} emerging as the most effective option. In particular, compared to the Zipf trace, the performance gap between \texttt{LRU} and \texttt{OPT} is significantly narrower. A possible explanation lies in the anticipated fluctuations in file popularity and the temporal correlations of requests, which can be advantageous to \texttt{LRU}. Additionally, both \texttt{D-NFPL} and \texttt{L-NFPL} perform worse than \texttt{S-NFPL}.

In the supplementary material (Section~\ref{app:additional_experiments}), we provide additional numerical results regarding the variability of the average miss ratio and running time for the experiments discussed in this section.

\subsection{Sampling in \texttt{NFPL}}
\label{ss:NFPL-Fix-Vs-NFPL-Var}

We evaluate the impact of different sampling methods within \texttt{S-NFPL} on the average miss ratio in the full observation setting, i.e., $p=1$. Specifically, we compare two sampling approaches: the first, which we call \texttt{\texttt{NFPL-Var}}, independently samples each request with probability~$q$, i.e., $(\beta_t)_{t=1}^{T}$ are i.i.d. Bernoulli random variables with parameter $q$. The second approach, which we refer to as \texttt{\texttt{NFPL-Fix}}, is slightly different. \texttt{\texttt{NFPL-Fix}} has a parameter $b$ such that for every batch of~$B$ requests, $b$ of them are selected uniformly at random, i.e., $(\beta_t)_{t=1}^{T}$ are Bernoulli random variables with parameter $b/B$ and $\sum_{t=kB+1}^{(k+1)B} \beta_t = b$, for any~$k$.

We compare the performance of \texttt{NFPL-Fix}, \texttt{NFPL-Var}, and \texttt{OPT} on all the considered traces. We consider a different variant of the Zipf-RR trace where are all items have the same total number of requests. We refer to this variant as Round-robin. The total number of requests in the Zipf, Akamai, and Round-robin is $5\times 10^{6}$, $2\times 10^{7}$, and $10^{6}$, respectively. Two cache sizes are considered in each trace: $C \in \{10, 200\}$ for the Zipf trace and $C \in \{10, 100\}$ for the Akamai and Round-robin trace.

Figure \ref{fig:NFPL-FIX-VAR} illustrates the average miss ratio for all the aforementioned caching policies when varying sampling probabilities, i.e., $q$ for \texttt{NFPL-Var} and $b/B$ for \texttt{NFPL-Fix}. 
  

Across the various traces we analyzed, the performance difference between \texttt{NFPL-Fix} and \texttt{NFPL-Var} is consistently minimal for all the sampling rates. This indicates that the selection of the sampling method may exert only a marginal impact on the performance of NFPL.

For the Zipf and Akamai traces, the performance of both \texttt{NFPL-Fix} and \texttt{NFPL-Var} tends towards that of \texttt{OPT} with increasing sampling rates. This is attributable to the relatively stationary nature of these traces, where the count of past requests serves as a good predictor for future requests; thus, more precise estimates bolster performance. In contrast, the round-robin trace benefits from noisier estimates, as it is preferable to overlook past requests in this scenario. As a result, the performance of \texttt{NFPL-Fix} and \texttt{NFPL-Var} deteriorates with a rising sampling rate.



%% file: Journal/conclusions.tex
In this paper, we studied the single cache problem under adversarial requests in the partial observability regime, where each request is observed with a specific probability. We proposed the \texttt{NFPL} caching policy, the first to achieve optimal regret bound with~$\mathcal{O}(1)$ amortized time complexity in this more restricted setting. Through experiments on both synthetic and real-world traces, we highlighted the practical importance of our theoretical results. In future work, we aim to explore no-regret caching policies for the more complex bipartite caching problem with partial observability of the requests, which arises in wireless content delivery networks.


%% file: Journal/ProofMainTh.tex




\subsection{Proof of Thm.~\ref{th:RegretNFPLMain}}
\label{Proof:Regret-NFPL}

The expected number of misses of any randomized caching policy~$\mathcal{A}$ is given by $\sum_{t=1}^{T} \Proba{f_t \notin S_{\mathcal{A}}(t)}$.
Note that for any \texttt{NFPL} variant, it can be shown that~$\boldsymbol{\gamma_t}$ is uniformly distributed in~$[0, \eta]^{N}$ for all~$t$, making the probability distribution of the set of stored items in the cache the same for any \texttt{NFPL} variant (see line~\ref{line:decison-rule} in Algorithm~\ref{algo:NFPL-singleCache}). Therefore, the expected number of misses remains the same, independent of the temporal correlations between these noise vectors.

For both \texttt{S-NFPL} and \texttt{D-NFPL}, it is evident that~$\boldsymbol{\gamma}_t$ is uniformly distributed in~$[0, \eta]^{N}$ for all~$t$. In the case of \texttt{L-NFPL}, the noise vector is inspired by the Follow-the-Lazy-Leader (\texttt{FLL}) algorithm from~\cite{kalai2005efficient}. The noise vector in~\eqref{e:Noise-FLL} can be rewritten as:
\begin{align}
\boldsymbol{\gamma}_t = \eta \left( \frac{\boldsymbol{\gamma}_0 - \hatbf{n}(t)}{\eta} - \ceil*{\frac{\boldsymbol{\gamma}_0 - \hatbf{n}(t)}{\eta}} \right).
\end{align}
Note that the random variable~$\mathbf{u} = \frac{\boldsymbol{\gamma}_0 - \hatbf{n}(t)}{\eta}$ is uniform within a range of width~$1$ in each component. Therefore,~$\mathbf{u} - \ceil*{\mathbf{u}}$ is a uniform random vector in~$[0,1]^N$, and finally,~$\boldsymbol{\gamma}_t$ is uniform in~$[0, \eta]^N$.

We first establish the regret bound for \texttt{NFPL} when~$pq=1$, before addressing the general scenario with arbitrary probabilities~$p$ and~$q$.

\vspace{5mm}

\noindent \textbf{Particular Case:} $pq=1$. Here, \texttt{NFPL} corresponds to \texttt{FPL} from~\cite{kalai2005efficient}, tailored for the caching problem in the full observation regime. This connection becomes clear when we cast the caching problem within the online linear learning framework outlined in Section~\ref{ss:OnlineLearning_FPL}, identifying the decision set~$\mathcal{X}$ and the cost set~$\mathcal{B}$ as introduced in that section. Finally, applying \cite[Thm. 1.1]{bhattacharjee2020fundamental}, we establish the regret bound for \texttt{NFPL}.

Since \texttt{NFPL} updates the cache state after every~$B$ requests, we model the caching problem as an online linear one, with a total number of rounds $T^{'}= \ceil*{T/B}$,
Let $t\in [T^{'}]$.

{\noindent \bf Cache state}. The set of cached items at any step~$t^{'}$, that satisfies $tB \leq t^{'} \leq (t+1)B -1$, is represented by the binary vector $\mathbf{x}_{t} = [x_{t,i}]_{i\in [N]}$, which indicates the files missing in the cache; that is, $x_{t,i} = 1$ if and only if file~$i$ is not stored in the cache at time~$t$, i.e., $i\notin S_{\mathcal{A}}(t^{'})$. A feasible cache allocation is then represented by a vector in the set:
\begin{equation}\label{eq:Capped-Simplex}
\mathcal{X} = \left \{ \mathbf{x} \in \{0,1\}^{N} \Bigg| \sum_{i=1}^N x_{i} = N-C \right \}.
\end{equation}

{\noindent \bf Requests}. The request process is represented as a sequence of vectors~$\mathbf{r}_t = (r_{t,i} \in \mathbb{N}: i \in [N])$ $\forall t$, where~$r_{t,f}$ is the number of requests received within the range $[(t-1)B+1, tB]$, i.e., $r_{t,f}=\sum_{s=(t-1)B+1}^{tB} \mathds{1}(f_s=f)$. Then, each of these vectors belongs to the set:

\begin{equation}\label{eq:requests_set}
\mathcal{B} = \left \{ \mathbf{r}\in  \mathbb{N}^N \Bigg| \;  \sum_{i=1}^N r_{i} = B  \right \}.
\end{equation}

{\noindent \bf Cost}. At each time slot~$t$, the cache pays a cost equal to the number of misses, i.e., to the number of requests for files not in the cache.
The cost can be computed as follows:
\begin{equation}\label{eq:cost}
 \scalar{\mathbf{r}_t}{\mathbf{x}_t}= \sum_{i=1}^N r_{t,i} x_{t,i}  .
\end{equation}

The decision vector of \texttt{FPL} with uniform noise~\cite{kalai2005efficient} in an online linear learning problem is as follows, 
    \begin{align} \label{e:helloAgain}
            \mathbf{x}_{t+1}(\texttt{FPL}) \in \argmin_{\mathbf{x}\in \mathcal{X}} \langle \mathbf{x}, \sum_{s=1}^{t} \mathbf{r}_s + \boldsymbol{\gamma}_{t+1} \rangle,
    \end{align}
where~$(\boldsymbol{\gamma}_t)_t$ are random vectors of~$N$ i.i.d. uniform random variables, each within the range~$[0,\eta]$. Let~$D$ be the diameter of the set~$\mathcal{X}$, i.e., $D=\sup_{\mathbf{x},\mathbf{y}\in \mathcal{X}} \norm{\mathbf{x}-\mathbf{y}}_1$, $A$ be a bound on the norm~$1$ of vectors in~$\mathcal{B}$, and~$R$ a bound on $\langle \mathbf{x},\mathbf{r} \rangle$ for any $(\mathbf{x},\mathbf{r})\in \mathcal{X}\times \mathcal{B}$.

The regret bound of \texttt{FPL} over $T^{'}$ rounds is upper bounded by $RAT^{'}/\eta+ D \eta$~\cite[Thm. 1.1]{kalai2005efficient}.

For the specific online learning problem with~$\mathcal{X}$ in~\eqref{eq:Capped-Simplex} and~$\mathcal{B}$ in~\eqref{eq:requests_set}, the vector~$\sum_{s=1}^{t} \mathbf{r_s}$ represents the number of requests for all files at step~$tB$. Moreover, the constants associated to the sets~$\mathcal{X}$ and~$\mathcal{B}$ are given by: $R=A=B$ and $D=2C$. Therefore, \texttt{FPL} in this setting stores the~$C$ files with the largest $n_f(tB)+\gamma_{tB,f}$ for $f\in \mathcal{I}$, which coincides with the update rule of \texttt{NFPL}, when $\eta=\sqrt{BT/2C}$, with $T$ the total number of requests. As a result, applying~\cite[Thm. 1.1]{kalai2005efficient} in this setting yields the following regret bound for \texttt{NFPL}:

\begin{align}\label{e:hello13-09}
      B^{2} T^{'}  \sqrt{\frac{2C}{BT} } &+ 2C    \sqrt{\frac{BT}{2C}}   \\
&=   \sqrt{2CB} \left( \frac{BT^{'}}{\sqrt{T}} + \sqrt{T}  \right)  \\
&\leq  \sqrt{2CB} \left ( B \left(\frac{T}{B} +1\right) \frac{1}{\sqrt{T}} + \sqrt{T} \right) \\ \label{e:hello13-09-blah}
&=  \sqrt{2CB} \left(2 \sqrt{T}  + \frac{B}{\sqrt{T}}\right).
\end{align}


\vspace{5mm}

\noindent \textbf{General Case: Arbitrary} $p$ \textbf{and} $q$. Note that to apply~\cite[Thm. 1.1 and Lem. 1.2]{kalai2005efficient}, \texttt{FPL} relies on the cumulative cost vector $\mathbf{r}_{1:t}$, which represents in the caching problem, the exact request count for each file. Unfortunately, \texttt{NFPL} with~$pq<1$ leverages approximate counts represented via the vector~$\hatbf{n}$ and therefore we can not use the results from~\cite{kalai2005efficient}. For this reason, we consider a variant of the online linear learning framework in Section~\ref{ss:OnlineLearning_FPL}, where at each step~$t$, the agent only observes a random vector~$\hatbf{r}_{t}$, which serves as an estimation of the exact cost vector~$\mathbf{r}_t$. We further assume that these random vectors satisfy Assumption~\ref{assum:NFPL-Regret}.
\begin{assumption}\label{assum:NFPL-Regret}
The vectors~$(\hatbf{r}_t)_{t=1}^{T^{'}}$ are bounded and independent such that: $\exists p\in \mathbb{R}^{+}\setminus \{0\}: \; \E{\hatbf{r}_t} = p \cdot \mathbf{r}_t$. 
\end{assumption}

We call \texttt{NFPL}, the algorithm that substitutes the exact cost vector by its noisy estimation within the \texttt{FPL} algorithm, i.e., 
\begin{align}\label{e:update-NFPL}
&\hatbf{x}_{t+1} = M(\hatbf{r}_{1:t}+ \boldsymbol{\gamma}_{t+1}), 
\end{align}
where $M(\mathbf{u})\in  \argmin_{\mathbf{x}\in \mathcal{X}} \langle \mathbf{u}, \mathbf{x} \rangle$, $\forall \mathbf{u}$.
Define the constants $\hat{A} = \sup_{\mathbf{\hat{r}} \in \mathcal{\hat{B}}} \norm{\frac{1}{p}\hatbf{r}}_1$ and $\hat{R} =~ \sup_{\mathbf{x} \in \mathcal{X}, \hatbf{r} \in \mathcal{\hat{B}}} \langle \frac{1}{p} \hatbf{r}, \mathbf{x} \rangle$, with $\mathcal{\hat{B}}$ being the set of all possible values of~$(\hatbf{r}_t)_{t=1}^{T^{'}}$.


Lemma~\ref{th:Regret-NFPL} shows that \texttt{NFPL}, with no assumption about the temporal correlations between the noise vectors $(\boldsymbol{\gamma^{'}}_t)_{t=1}^{T^{'}}$, retains the~$\mathcal{O}(\sqrt{T^{'}})$ regret guarantees. 



\begin{lemmaSp}\label{th:Regret-NFPL} Under Assumption~\ref{assum:NFPL-Regret}:

\begin{align}
\mathcal{R}_{T^{'}}(\texttt{NFPL}) \leq \frac{p}{\eta} \hat R \hat A T^{'} + \frac{\eta}{p} D.
\end{align}
\end{lemmaSp}



\begin{IEEEproof}[Proof of Lemma~\ref{th:Regret-NFPL}]
To relate \texttt{NFPL} to \texttt{FPL}, we consider a fictitious online linear problem with the same decision set as the original one, namely~$\mathcal{X}$. On the other hand, the costs set, denoted~$\hat{\mathcal{D}}$, depends on the set of all possible values for the random vectors $(\hatbf{r}_t)_t$ as follows,  
\begin{align}
        \hat{\mathcal{D}} = \left \{ \frac{1}{p} \cdot \mathbf{a}: \; \mathbf{a}\in \hat{\mathcal{B}} \right \}. 
\end{align}
The analogue of the constants~$R$ and~$A$ in the original problem are~$\hat{R}$ and~$\hat{A}$. Therefore, if the parameter of the noise vectors used is $\eta^{'}= \eta /p$, then using \cite[Theorem 1.1 a)]{kalai2005efficient}, the regret of \texttt{FPL} in the fictitious problem, namely~$\hat{\mathcal{R}}_{T^{'}}(\texttt{FPL})$, is bounded by~$\frac{p}{\eta} \hat R \hat A T^{'} + \frac{\eta}{p} D$.

To prove that~$\mathcal{R}_T(\texttt{NFPL})\leq \hat{\mathcal{R}}_T(\texttt{FPL})$, we consider an adversary that selects~$\frac{1}{p}\hatbf{r}_t$. It follows that the update rule~$\mathbf{d}_{t}$ of \texttt{FPL} in this case is given by, 
\begin{align}
        \mathbf{d}_{t+1} = M( \hatbf{r}_{1:t}/ p  + \boldsymbol{\gamma}_{t+1}/ p) &= M(\hatbf{r}_{1:t} + \boldsymbol{\gamma}_{t+1} )\\
        &= \hatbf{x}_{t+1},
\end{align}  
where $\hatbf{r}_{1:t}= \sum_{s=1}^{t}\mathbf{r}_s$. Let~$G$ represent the cost of the \texttt{FPL} algorithm when the adversary selects $\frac{1}{p}\hatbf{r}_t$; formally, $G = \sum_{t=1}^{T^{'}} \left\langle \frac{\hat{\mathbf{r}}_t}{p}, \mathbf{d}_t \right\rangle$. 
Additionally, let $G^{*}$ denote the corresponding static optimum, given by $G^{*} = \left\langle \frac{\hatbf{r}_{1:T^{'}}}{p}, M\left( \frac{\hatbf{r}_{1:T^{'}}}{p} \right) \right\rangle$.

We further define $H$ as the cost of \texttt{NFPL} in the original problem when the adversary selects the sequence $(\mathbf{r}_t)_{t=1}^{T^{'}}$, given by 
$H = \sum_{t=1}^{T^{'}} \left\langle \mathbf{r}_t, \hat{\mathbf{x}}_t \right\rangle$. Similarly, let $H^{*}$ be the corresponding static optimum, defined as $H^{*} = \left\langle \mathbf{r}_{1:T^{'}}, M(\mathbf{r}_{1:T^{'}}) \right\rangle$. Therefore, we can write: 
    \begin{align*}
\hat{\mathcal{R}}_{T^{'}}(\texttt{FPL}) -  \E{(H - H^{*})}
&\geq \E{G- H} + \E{H^{*}- G^{*}} 
    \end{align*}

Observe that~$\hatbf{x}_t$ depends only on $(\hatbf{r}_s)_{s=1}^{t-1}$ and hence it is independent from~$\hatbf{r}_t/p$, thanks to Assumption~\ref{assum:NFPL-Regret}. Therefore, 
\begin{align*}
\E{G}= \sum_{t=1}^{T^{'}} \scalar{\E{\hatbf{r}_t/p}}{\E{\mathbf{d}_t}} = \E{\sum_{t=1}^{T^{'}} \scalar{\mathbf{r}_t}{\hatbf{x}_{t}}}= \E{H}
\end{align*}
Moreover, 
\begin{align*}
\E{G^{*}}
&= \E{\left\langle \hatbf{r}_{1:T^{'}} / p ,M(\hatbf{r}_{1:T^{'}}/p) \right\rangle}\\
&\leq \E{\scalar{\hatbf{r}_{1:T^{'}}/p}{M(\mathbf{r}_{1:T^{'}})}}\\
&= \scalar{\mathbf{r}_{1:T^{'}}}{M(\mathbf{r}_{1:T^{'}})} = H^{*}
\end{align*}

Finally, since $\E{G-H}=0$ and $\E{H^{*}- G^{*}}\geq 0$, we conclude that for any sequence~$(\mathbf{r}_t)_{t=1}^{T^{'}}$, $\E{H - H^{*}}\leq \hat{\mathcal{R}}_{T^{'}}(\texttt{FPL})\leq \frac{p}{\eta} \hat R \hat A T^{'} + \frac{\eta}{p} D$. This finishes the proof.

\end{IEEEproof}

To recover the update rule of the caching policy \texttt{NFPL}, consider the general \texttt{NFPL} applied for the online linear learning problem with~$\mathcal{X}$ in~\eqref{eq:Capped-Simplex}, $\mathcal{B}$ in~\eqref{eq:requests_set}, and~$\hatbf{r}_t=(\hat r_{t,f})_{f\in \mathcal{I}}$ set to, 
 \begin{align}
            \hatbf{r}_t = \sum_{s=(t-1)B+1}^{tB} \delta_s \beta_s\mathds{1}(f_s=f).
\end{align}
It follows that $\E{\hatbf{r}_t} = pq \mathbf{r}_t$ and therefore $\hat R= \hat A = B/pq$, and $D=2C$. Thanks to Lemma~\ref{th:Regret-NFPL}, we obtain the desired regret bound following similar steps as in~\eqref{e:hello13-09}-\eqref{e:hello13-09-blah}. This finishes the proof of Theorem~\ref{th:RegretNFPLMain}.


\subsection{Proof of Thm.~\ref{th:TimeComplexityNFPL}}
\label{Proof:TimeComplexityNFPL}

Regarding the implementations of the three variants of \texttt{NFPL}: \texttt{S-NFPL}, \texttt{D-NFPL}, and \texttt{L-NFPL}, we examine for each first the case where~$B=1$ and~$pq=1$, before moving to the general case. The most computationally demanding operation in the algorithm is identifying the top~$C$ files in terms of perturbed counts, $\hatbf{n}+\boldsymbol{\gamma}$.

\vspace{2mm}
\noindent  \textbf{D-NFPL}. At each time step, the noise vectors~$(\boldsymbol{\gamma}_t)_{t\in [T]}$ are regenerated. Consequently, all entries of~$\hat{\mathbf{n}} + \boldsymbol{\gamma}$ are updated, necessitating a sorting operation at every time step to find the top~$C$ files. This leads to an amortized time complexity of~$\mathcal{O}(N\log N)$. When~$B>1$, sorting occurs only when~$t$ is a multiple of~$B$, reducing the time complexity by a factor of~$\frac{1}{B}$.


\vspace{2mm}

\noindent  \textbf{S-NFPL}. The noise vectors remain constant over time, so only the component corresponding to the requested item in $\hat{\mathbf{n}} + \boldsymbol{\gamma}$ is updated at each time step. This allows us to utilize a \textit{Heap} data structure to efficiently keep track of the top~\(C\) files. Specifically, at each step \(t\), if \(f_t \in S_{t-1}\), the corresponding count in the heap is incremented by~1, while maintaining the heap structure with a time complexity of~\(\mathcal{O}(\ln{C})\). If \(f_t\) is not in \(S_{t-1}\), the file with the smallest perturbed count is retrieved from the heap in~\(\mathcal{O}(1)\) time. This file’s count is then compared to~\(\hat{n}_{f_t}(t) + \gamma_{t,f_t}\). If the perturbed count of~\(f_t\) exceeds that of the smallest file in the heap, the smallest file is removed, and \(f_t\) is inserted into the heap with time complexity of~$\mathcal{O}(\ln{C})$. This operation maintains the heap structure and ensures that the heap always contains the top~\(C\) files based on their perturbed counts. It becomes clear now that the amortized time complexity of \texttt{S-NFPL} is~$\mathcal{O}(\ln{C})$. If~$pq<1$, no updates are made to the heap data structure whenever the request is not taken into account, i.e., $\delta_t \beta_t=0$, which happens with probability~$pq$, justifying the~$\mathcal{O}(pq\ln{C})$ amortized time complexity. Whether or not~$B$ is greater than~$1$, the heap updates must occur whenever~$\delta_t\beta_t=1$, which explains the exclusion of~$B$ from the time complexity. 

\vspace{2mm}

\noindent \textbf{L-NFPL}. Define the vector~$\mathbf{m}_t$ as follows, $\mathbf{m}_t = \hat{\mathbf{n}}_t + \boldsymbol{\gamma}_t$. Given the expression of the noise vector in \texttt{L-NFPL} in~\eqref{e:Noise-FLL}, $\gamma_{t,f}- \gamma_{t-1,f}>0$ if and only if~$f=f_t$. Therefore, the vector~$\mathbf{m}$ has the property that at most one of its components is updated at each step, which allows to employ a heap to keep track of the top~$C$ files and to incur a time complexity of~$\mathcal{O}(\ln{C})$ per insertion/deletion, similarly to \texttt{S-NFPL}. However, in \texttt{L-NFPL}, it is possible to have $\mathbf{m}_t=\mathbf{m}_{t-1}$ when $\gamma_{t,f_t}=\gamma_{t-1,f_t}- 1$. Therefore, whenever $\gamma_{t,f_t}=\gamma_{t-1,f_t}- 1$, the counts of all items remain the same, alleviating the need for updating the heap data structure, and the time complexity in these steps is~$\mathcal{O}(1)$. The amortized time complexity is then proportional to: 
\begin{align}
    \frac{\ln C}{T}\sum_{t=1}^{T}\Proba{\gamma_{t,f_t}\neq \gamma_{t-1,f_t}- 1}.  
\end{align}
Next, we prove that~$\Proba{\gamma_{t,f_t}\neq \gamma_{t-1,f_t}- 1}\leq \frac{pq}{\eta}$, justifying the amortized time complexity of \texttt{L-NFPL} since~$\eta = \sqrt{BT/2C}$. 

\noindent Define~$\Delta_t$ as~$\ceil*{\frac{\hat n_{f_t}(t)-\gamma_{0,f_t}}{\eta}}-\ceil*{\frac{\hat n_{f_t}(t-1)-\gamma_{0,f_t}}{\eta}}$. 
\begin{align}
    \Proba{\gamma_{t,f_t}- \gamma_{t-1,f_t}+ 1\neq 0} 
&= \Proba{\eta \Delta_t \neq 0} \\ 
&= \Proba{\Delta_t\geq 1},
\end{align}
where the last step follows because~$\Delta_t$ is a positive integer. 

To compute $\Proba{\Delta_t\geq 1}$, observe that if~$\delta_t\beta_t=0$, $\hat n_{f_t}(t) = \hat n_{f_t}(t-1)$ and hence $\Delta_t=0$, which enables us to write:  
\begin{align} \nonumber
       & \Proba{\Delta_t \geq 1}  \\ \nonumber
    &= \Proba{\Delta_t \geq 1 | \delta_t\beta_t=1 } pq + \Proba{\Delta_t \geq 1 | \delta_t\beta_t=0 } (1-pq) \\ \label{e:helloworld}
    &= \Proba{\Delta_t \geq 1 | \delta_t\beta_t=1 }pq.
\end{align}

If $\eta \leq  1$, the statement~$\Proba{\Delta_t\geq 1}\leq pq/\eta$ becomes evident thanks to~\eqref{e:helloworld}.

We assume that $\eta> 1$ and $\delta_t\beta_t= 1$. Define the following quantities:
\begin{align}
      z = \frac{\hat n_{f_t}(t) -\gamma_{0,f_t}}{\eta}, \; k  =  \ceil*{z}.
\end{align} 

A direct consequence of $\delta_t\beta_t= 1$ is $\hat n_{f_t}(t)= \hat n_{f_t}(t-1)+1$, we can then write:
\begin{align} \nonumber
        &\Delta_t = k- \ceil*{z-\frac{1}{\eta}}  \leq z +1  - z +\frac{1}{\eta} = 1 + \frac{1}{\eta} <2\\ \label{e:helloworld2}
        &\implies \Delta_t \leq 1.  
\end{align}
Moreover, 
\begin{align}\nonumber
        \Delta_t = 1 
&\iff  k = \ceil*{z-\frac{1}{\eta}} + 1 \\ \nonumber
&\iff   k -2  <z -  \frac{1}{\eta} \leq k-1\\ \label{e:helloworld3}
& \iff -2 + \frac{1}{\eta}  <  z-k \leq -1 +  \frac{1}{\eta}. 
\end{align}
Combining \eqref{e:helloworld},\eqref{e:helloworld2}, and \eqref{e:helloworld3}, we obtain:
\begin{align}
\Proba{\Delta_t \geq 1} 
&= \Proba{\Delta_t \geq 1 | \delta_t\beta_t=1 } pq \\
&=  \Proba{\Delta_t = 1 | \delta_t\beta_t=1 } pq \\ \label{e:mon19_1}
&= \Proba{-2 + \frac{1}{\eta}  <  z-k  \leq -1 + \frac{1}{\eta}} pq \\ \label{e:mon19_2}
&= \Proba{-1  <  z-k  \leq -1 + \frac{1}{\eta}}pq \\ \label{e:mon19_3}
&\leq \frac{pq}{\eta}. 
\end{align}
where the transitions \eqref{e:mon19_1}-\eqref{e:mon19_2} and \eqref{e:mon19_2}-\eqref{e:mon19_3} used the fact that~$z-k$ is uniform over~$[-1,0]$ for any value of~$n_{f_t}$. The correctness of the statement above is due to~$z$ being uniformly distributed over an interval of width equal to $1$. Indeed, since~$\gamma_{0,f_t}$ is uniform over~$[0,\eta]$, it follows that,~$z$ is uniform over~$[\frac{n_{f_t}}{\eta} -1 , \frac{n_{f_t}}{\eta}]$.

Similarly to \texttt{S-NFPL}, even if~$t$ is a multiple of~$B>1$, the update of the heap is necessary whenever~$\delta_t\beta_t=1$, explaining the absence of the factor~$1/B$ from the amortized time complexity. This completes the proof.

\subsection{Additional Experimental Results}
\label{app:additional_experiments}

In this section, we report more detailed results on the comparison among the variants of \texttt{NFPL}, \texttt{LFU}, and \texttt{LRU} in terms of Average miss ratio, variance, and execution time, for each trace and different sampling probabilities. In particular, Tables~\ref{tab:zipf1}-\ref{tab:zipf0001} show the results on the Zipf distribution. The results for the Zipf-RR are shown in Tables~\ref{tab:zipfRR1}-\ref{tab:zipfRR0001}. Finally, Tables~\ref{tab:aka1}-\ref{tab:aka0001} show the results regarding the Akamai trace.

\begin{table}[h]
    \caption{Zipf with $p=1$}
    \label{tab:zipf1}
    \centering
    \scriptsize
    \begin{tabular}{l c c c}
        \toprule
        Algorithm & Average miss ratio  & Variance &  Execution Time (sec)  \\
        \midrule
        \texttt{L-NFPL} & $0.49$ & $1.27\times10^{-4}$ & $3.77\times10^{-1}$ \\
        \texttt{S-NFPL}  & $0.48$ & $1.36\times10^{-4}$ & $4.26\times10^{-1}$ \\
        \texttt{D-NFPL} & $0.48$ & $6.17\times10^{-5}$ & $5.48$ \\
        \texttt{LFU} & $\mathbf{0.47}$ & $\mathbf{1.11\times10^{-6}}$ & $\mathbf{3.5\times10^{-1}}$ \\
        \texttt{LRU} & $0.61$ & $0.49\times10^{-7}$ & $9.82\times10^{-1}$ \\
        \bottomrule
    \end{tabular}
\bigskip
\caption{Zipf with $p=0.7$}
    \label{tab:zipf07}
    \centering
    \scriptsize
    \begin{tabular}{l c c c}
        \toprule
        Algorithm & Average miss ratio  & Variance &  Execution Time (sec)  \\
        \midrule
        \texttt{L-NFPL} & $0.49$ & $1.9\times10^{-4}$ & $2.85\times10^{-1}$ \\
        \texttt{S-NFPL}  & $0.48$ & $2.23\times10^{-4}$ & $3.11\times10^{-1}$ \\
        \texttt{D-NFPL} & $0.48$ & $1.33\times10^{-4}$ & $1.2$ \\
        \texttt{LFU} & $\mathbf{0.47}$ & $\mathbf{7.32\times10^{-6}}$ & $\mathbf{2.6\times10^{-1}}$ \\
        \texttt{LRU} & $0.61$ & $2.86\times10^{-4}$ & $7.2\times10^{-1}$ \\
        \bottomrule
    \end{tabular}
\bigskip
\caption{Zipf with $p=0.01$}
    \label{tab:zipf0001}
    \centering
    \scriptsize
    \begin{tabular}{l c c c}
        \toprule
        Algorithm & Average miss ratio  & Variance &  Execution Time (sec)  \\
        \midrule
        \texttt{L-NFPL} & $0.51$ & $1.56\times10^{-2}$ & $2.81\times10^{-1}$ \\
        \texttt{S-NFPL}  & $0.51$ & $1.56\times10^{-2}$ & $2.54\times10^{-1}$ \\
        \texttt{D-NFPL} & $\mathbf{0.5}$ & $\mathbf{1.58\times10^{-4}}$ & $1.03$ \\
        \texttt{LFU} & $0.51$ & $1.54\times10^{-2}$ & $\mathbf{2.46\times10^{-1}}$ \\
        \texttt{LRU} & $0.62$ & $1.11\times10^{-2}$ & $2.78\times10^{-1}$ \\
        \bottomrule
    \end{tabular}
\end{table}
\textbf{Zipf}. In Tables~\ref{tab:zipf1}-\ref{tab:zipf0001} we can observe that \texttt{L-NFPL} achieves the lowest execution time among all the variants of \texttt{NFPL}, except for $p=0.01$ (where \texttt{S-NFPL} has the lowest execution time). Among all the \texttt{NFPL} policies, \texttt{D-NFPL} presents the lowest variance, in line with the discussion in Section~\ref{ss:Tradeoff_regret_time}. In such a static scenario, \texttt{LFU} performs optimally with the smallest average miss ratio. 

\begin{table}[t]
    \caption{Zipf-RR with $p=1$}
    \label{tab:zipfRR1}
    \centering
    \scriptsize
    \begin{tabular}{l c c c}
        \toprule
        Algorithm & Average miss ratio  & Variance &  Execution Time (sec)  \\
        \midrule
        \texttt{L-NFPL} & $0.48$ & $1.7\times10^{-4}$ & $3.69\times10^{-1}$ \\
        \texttt{S-NFPL}  & $0.49$ & $1.12\times10^{-4}$ & $4.23\times10^{-1}$ \\
        \texttt{D-NFPL} & $\mathbf{0.48}$ & $1.09\times10^{-4}$ & $3.11$ \\
        \texttt{LFU} & $0.57$ & $\mathbf{1.07\times10^{-6}}$ & $\mathbf{3.63\times10^{-1}}$ \\
        \texttt{LRU} & $0.57$ & $1.11\times10^{-6}$ & $9.15\times10^{-1}$ \\
        \bottomrule
    \end{tabular}
\bigskip
\caption{Zipf-RR with $p=0.7$}
    \label{tab:zipfRR07}
    \centering
    \scriptsize
    \begin{tabular}{l c c c}
        \toprule
        Algorithm & Average miss ratio  & Variance &  Execution Time (sec)  \\
        \midrule
        \texttt{L-NFPL} & $0.49$ & $2.32\times10^{-4}$ & $2.77\times10^{-1}$ \\
        \texttt{S-NFPL}  & $0.49$ & $1.26\times10^{-4}$ & $3.13\times10^{-1}$ \\
        \texttt{D-NFPL} & $\mathbf{0.48}$ & $1.27\times10^{-4}$ & $1.22$ \\
        \texttt{LFU} & $0.5$ & $1.48\times10^{-4}$ & $\mathbf{2.6\times10^{-1}}$ \\
        \texttt{LRU} & $0.54$ & $2.26\times10^{-4}$ & $6.64\times10^{-1}$ \\
        \bottomrule
    \end{tabular}
\bigskip
\caption{Zipf-RR with $p=0.01$}
    \label{tab:zipfRR0001}
    \centering
    \scriptsize
    \begin{tabular}{l c c c}
        \toprule
        Algorithm & Average miss ratio  & Variance &  Execution Time (sec)  \\
        \midrule
        \texttt{L-NFPL} & $0.51$ & $6.32\times10^{-4}$ & $2.71\times10^{-1}$ \\
        \texttt{S-NFPL}  & $\mathbf{0.5}$ & $6.36\times10^{-4}$ & $2.47\times10^{-1}$ \\
        \texttt{D-NFPL} & $0.51$ & $7.98\times10^{-4}$ & $1.04$ \\
        \texttt{LFU} & $\mathbf{0.5}$ & $\mathbf{4.64\times10^{-4}}$ & $\mathbf{2.38\times10^{-1}}$ \\
        \texttt{LRU} & $0.48$ & $1.69\times10^{-4}$ & $2.67\times10^{-1}$ \\
        \bottomrule
    \end{tabular}
\end{table}

\textbf{Zipf-RR}. As already discussed in Section~\ref{sec:experiments}, with the Zipf-RR traces there is a significant difference concerning the Zipf, especially with $p=1$ (Table~\ref{tab:zipfRR1}), where the best average miss ratio is achieved by \texttt{D-NFPL} and \texttt{L-NFPL}. When $p=1$, \texttt{D-NFPL} presents the smallest variance and largest execution time among the variants of \texttt{NFPL}. Such an execution time is due to the sorting performed by \texttt{D-NFPL} at each step. Similar to the Zipf, \texttt{L-NFPL} has the best execution time among all the \texttt{NFPL} policies except for the case when the sampling probability is low (Table~\ref{tab:zipfRR0001}). In such a case, \texttt{S-NFPL} achieves the lowest execution time given that having a low sampling probability is almost equivalent to having a smaller number of requests.
\begin{table}[t]
    \caption{Akamai with $p=1$}
    \label{tab:aka1}
    \centering
    \scriptsize
    \begin{tabular}{l c c c}
        \toprule
        Algorithm & Average miss ratio  & Variance &  Execution Time (sec)  \\
        \midrule
        \texttt{L-NFPL} & $0.2$ & $2.08\times10^{-4}$ & $3.77\times10^{-1}$ \\
        \texttt{S-NFPL}  & $\mathbf{0.19}$ & $1.7\times10^{-4}$ & $2.65\times10^{-1}$ \\
        \texttt{D-NFPL} & $0.2$ & $7.82\times10^{-5}$ & $2.61\times 10^{-1}$ \\
        \texttt{LFU} & $\mathbf{0.19}$ & $\mathbf{4.52\times10^{-5}}$ & $\mathbf{2.2\times10^{-1}}$ \\
        \texttt{LRU} & $0.22$ & $3.35\times10^{-6}$ & $1.17$ \\
        \bottomrule
    \end{tabular}
\bigskip
\caption{Akamai with $p=0.7$}
    \label{tab:aka07}
    \centering
    \scriptsize
    \begin{tabular}{l c c c}
        \toprule
        Algorithm & Average miss ratio  & Variance &  Execution Time (sec)  \\
        \midrule
        \texttt{L-NFPL} & $0.2$ & $2.1\times10^{-4}$ & $2.82\times10^{-1}$ \\
        \texttt{S-NFPL}  & $0.2$ & $1.86\times10^{-4}$ & $2.04\times10^{-1}$ \\
        \texttt{D-NFPL} & $0.2$ & $1.02\times10^{-4}$ & $1.99\times 10^{-1}$ \\
        \texttt{LFU} & $\mathbf{0.19}$ & $1.16\times10^{-4}$ & $\mathbf{1.54\times10^{-1}}$ \\
        \texttt{LRU} & $0.22$ & $\mathbf{8.96\times10^{-5}}$ & $8.37\times10^{-1}$ \\
        \bottomrule
    \end{tabular}
\bigskip

\caption{Akamai with $p=0.01$}
    \label{tab:aka0001}
    \centering
    \scriptsize
    \begin{tabular}{l c c c}
        \toprule
        Algorithm & Average miss ratio  & Variance &  Execution Time (sec)  \\
        \midrule
        \texttt{L-NFPL} & $0.144$ & $3.1\times10^{-4}$ & $4.98\times10^{-1}$ \\
        \texttt{S-NFPL}  & $0.144$ & $3.1\times10^{-4}$ & $4.63\times10^{-1}$ \\
        \texttt{D-NFPL} & $\mathbf{0.142}$ & $3.5\times10^{-4}$ & $5.58\times 10^{-1}$ \\
        \texttt{LFU} & $0.144$ & $\mathbf{3\times10^{-4}}$ & $\mathbf{4.56\times10^{-1}}$ \\
        \texttt{LRU} & $0.16$ & $2.14\times10^{-4}$ & $5.64\times10^{-1}$ \\
        \bottomrule
    \end{tabular}
\end{table}

\textbf{Akamai}. With the real trace of Akamai, all the compared solutions are very close in terms of average miss ratio (except for \texttt{LRU} with $p=0.01$). Surprisingly, in this scenario, we can observe an inverted trend concerning the two cases before: the average miss ratio decreases as the probability of sampling decreases as well. \texttt{N\_FPL} does not show the best average miss ratio since, with $N=1000$ and $C=100$, the algorithm has not reached the asymptotic regime yet.